\documentclass{article}
\usepackage[preprint]{neurips_2022}
\usepackage[hidelinks]{hyperref}       
\usepackage{amsmath} 
\usepackage{amsfonts}       
\usepackage{booktabs}       

\usepackage[draft,layout=margin]{fixme}
\usepackage{bm} 
\usepackage{placeins}
\usepackage{caption}
\usepackage{textcomp}
\usepackage{geometry} 
\usepackage{tikz}
\usepackage{eurosym}
\newcommand{\mybox}[4]{
    \begin{figure}[!ht]
        \centering
    \begin{tikzpicture}
        \node[anchor=text,text width=\columnwidth-1.2cm, draw, rounded corners, line width=1pt, fill=#3, inner sep=5mm, align=justify] (big) {\color{black}\\#4};
        \node[draw, rounded corners, line width=.5pt, fill=#2, anchor=west, xshift=5mm] (small) at (big.north west) {\color{black}#1};
    \end{tikzpicture}
    \end{figure}
}
\makeatletter
\renewcommand{\paragraph}[1]{\medskip\noindent\textbf{#1}\quad}
\makeatother

\begin{document}
\let\WriteBookmarks\relax
\def\floatpagepagefraction{1}
\def\textpagefraction{.001}



\title{Applied metamodelling for ATM performance simulations}                      



%

\author{%
  Christoffer Riis\\
  DTU Management\\
  Technical University of Denmark\\
  \texttt{chrrii@dtu.dk} \\
  \And
  Francisco N. Antunes\\
  DTU Management\\
  Technical University of Denmark\\
  \texttt{franant@dtu.dk} \\
  \And
  Tatjana Boli\'c\\
  School of Architecture and Cities \\
  University of Westminster\\
  \texttt{t.bolic@westminster.ac.uk} \\
  \And
  Gérald Gurtner\\
  School of Architecture and Cities \\
  University of Westminster\\
  \texttt{g.gurtner@westminster.ac.uk} \\
  \And
  Andrew Cook\\
  School of Architecture and Cities \\
  University of Westminster\\
  \texttt{cookaj@westminster.ac.uk} \\
  \And
  Carlos Lima Azevedo\\
  DTU Management\\
  Technical University of Denmark\\
  \texttt{climaz@dtu.dk} \\
  \And
  Francisco Câmara Pereira\\
  DTU Management\\
  Technical University of Denmark\\
  \texttt{camara@dtu.dk} \\
}

\maketitle

\begin{abstract}
The use of Air traffic management (ATM) simulators for planing and operations can be challenging due to their modelling complexity. This paper presents XALM (eXplainable Active Learning Metamodel), a three-step framework integrating active learning and SHAP (SHapley Additive exPlanations) values into simulation metamodels for supporting ATM decision-making. XALM efficiently uncovers hidden relationships among input and output variables in ATM simulators, those usually of interest in policy analysis. Our experiments show XALM's predictive performance comparable to the XGBoost metamodel with fewer simulations. Additionally, XALM exhibits superior explanatory capabilities compared to non-active learning metamodels.

Using the `Mercury' (flight and passenger) ATM simulator, XALM is applied to a real-world scenario in Paris Charles de Gaulle airport, extending an arrival manager's range and scope by analysing six variables. This case study illustrates XALM's effectiveness in enhancing simulation interpretability and understanding variable interactions. By addressing computational challenges and improving explainability, XALM complements traditional simulation-based analyses.

Lastly, we discuss two practical approaches for reducing the computational burden of the metamodelling further: we introduce a stopping criterion for active learning based on the inherent uncertainty of the metamodel, and we show how the simulations used for the metamodel can be reused across key performance indicators, thus decreasing the overall number of simulations needed. 
\end{abstract}

\section{Introduction}
Accurately modelling modern air traffic management (ATM) systems is challenging due to their complexity. ATM systems involve a wide range of stakeholders, variables, uncertainty, and human behaviour, which interact across both airspace and ground operations levels~\citep{cook2016complexity}. Ensuring the safety and efficiency of air traffic flows, while taking these factors into account, makes ATM systems difficult to model~\citep{cook2007european}. However, fast-time simulation approaches provide a way to study and model ATM systems \citep{riis2021active,raquel2022rnest}. These approaches allow researchers and practitioners to explore, test, and propose designs and solutions that would be impractical to implement in the real world. Simulators provide a virtual environment in which these studies can be conducted. Technically, a `simulator' is the computer implementation of a conceptual `simulation model', but for simplicity, we use both terms interchangeably in this paper.

Despite the benefits of using simulation-based solutions for modelling complex and real-world stochastic systems and their evolution over time, these approaches have a major drawback: they end up being tools with limited understandability, transparency, and interpretability regarding the details of the underlying simulation model and the results obtained for the non-expert end-users and policymakers. This is largely a consequence of the nature of simulation modelling rather than a deliberate design choice made by the developers of simulators. While it is expected that the internal components of the simulators (such as equations, functions, and theoretical foundations) and their logical interactions can be clearly identified and understood, the external, emergent behaviour of simulators may be less explainable. This lack of understandability can be particularly challenging for end-users, policymakers, and non-experts who are not familiar with the intricate implementation details thereof.\\

\textbf{ATM performance assessment in Europe}\quad

An important element of SESAR\footnote{\url{https://www.sesarju.eu}}, as the technological pillar of the Single European Sky initiative, is to bring about improvements, as measured through specific key performance indicators (KPIs), and as implemented by a series of so-called SESAR `Solutions'. SESAR `Solutions' represent a change in the way air traffic management is performed and are new operational concepts, procedures, and relevant technologies\footnote {A recently-launched `digital catalogue' is available at \url{https://www.sesarju.eu/catalogue}.}.

Central to performance assessment in SESAR is its Performance Framework. This is partly supported by EATMA – the European Air Traffic Management Architecture, which is the common architecture framework for SESAR, and the means of integrating operational and technical content developments. The term `metamodel' is not used extensively in these various SESAR contexts, and typically describes, at a high level, logical entity relationships, e.g. for performance data and as an architecture mapping
and database model. 

Different SESAR Solutions variously deploy different simulations to demonstrate their expected performance contributions across the International Civil Aviation Organisation (ICAO) set of eleven key performance areas (KPAs). They use a number of specific KPIs defined in the Performance Framework. The Solutions are, indeed, compelled to assess performance expectations as part of the SESAR programme. This brings challenges in terms of computational effort, simulation consistency, assessing KPI interdependencies and general integration.

Simulation-based studies in ATM often thus focus on assessing the performance impact of proposed solutions and concepts~\citep{sesarju2018,EUmasterplan2020,bolic2021sesar}, on existing or planned systems, usually focusing on a single solution. Simulation studies often rely on manual exploration of the underlying simulator's behaviour, using domain knowledge, expert-driven scenario design, and `what-if' approaches to reduce and discretise the input space into a limited set of possible system states to investigate. While these methodologies can be useful for limiting the space of investigation and communicating results to stakeholders, there is a risk of failing to properly assess the behaviour of the simulated system in its entirety, leaving relevant simulation input regions unexplored. As a result, the simulation analysis is restricted to small regions of the uncertainty space, limiting the insights and conclusions that can be drawn from the simulation. To this end, particularly meaningful exploration of new scenarios through complex simulators is often impractical and expensive in the effort required for purposeful analyses of results.\\

\textbf{Exploration of simulators}\quad
The ATM research and development community is thus well-acquainted with the trade-offs between the advantages and disadvantages of fast-time simulation modelling, as they have a long history of using such techniques~\citep{phillips2000validation,cook2007european,eurocontrol2010european,GURTNER201726,delgado2021network}. While simulation models are typically simplified representations of real systems~\citep{law2015simulation}, they can still be complex software programs with many input and output variables and parameters, requiring a large amount of data and computational resources for calibration. The complexity of a simulation model increases with the degree of detail, realism, and purpose of the simulation, as well as the complexity of the system being modelled and the problem being addressed. The large dimensional sizes and value ranges of input spaces can be a significant burden when exploring the behaviour of the simulation as a whole. In addition, the lack of an explicit and manageable closed-form function, as opposed to pure analytical approaches, can make it difficult to understand the true impact of input variables and parameters on the system's output metrics or KPIs and their interrelationships.

To realise the full potential of advanced ATM simulators, the researcher must overcome the computational burden of using the simulator and the subsequent results analyses. However, in this field, and literature, there is a tendency to push the complexity to the limit, neglecting the actual computational cost of extensively using the simulator afterwards. The increase in computer power has just increased the complexity of the simulator, and thus the computational time of running the simulators is still very high. One way to reduce the computational burden of using advanced ATM simulators is to combine them with other methods, such as \textit{metamodels}. These methods can interpolate the simulator's outputs as well as the associated uncertainty and variance, thus enabling the researcher to effectively explore the scenario space of the simulator without being limited by the computational cost of running the simulator extensively. 
Therefore, metamodels are an important tool for decision and policymakers to efficiently utilise advanced air traffic management simulators.\\

\textbf{Metamodels}\quad
Metamodelling is a framework for approximating the unknown relationship between the inputs and outputs of a simulation model with an explicit, known functional form, such as a statistical or machine learning model~\citep{friedman2012simulation}. We focus on fast-time simulation metamodels, which are specifically designed to reproduce the behaviour of simulation models~\citep{kleijnen2000methodology,friedman2012simulation,gramacy2020surrogates}.

\begin{figure}[t]
    \centering
    \includegraphics[width=0.70\textwidth]{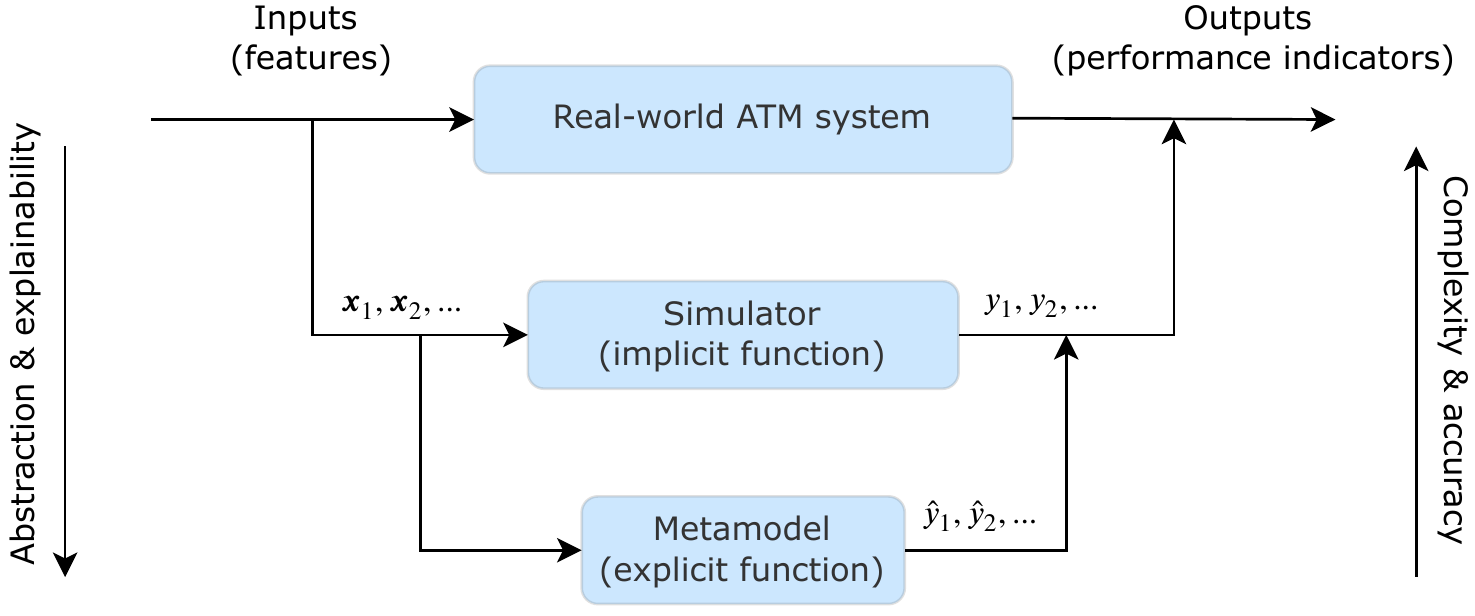}
    \caption{Relationship between the real-world system under study, the simulator, and the metamodel.}
    \label{fig:metamodel}
\end{figure}

A simulation metamodel (for simplicity, `metamodel') can be seen as an abstraction of a simulator, in the same way that a simulator represents an abstraction of a real-world system or phenomenon, as illustrated in~\autoref{fig:metamodel}. 
The main advantage of metamodels is that they can reduce the computational cost of running time-consuming simulation experiments by exploiting their approximate nature, functional simplicity, and fast computing. Since they approximate the underlying simulation functions, metamodels can achieve a balance between computational speed and controlled accuracy loss, depending on their goals. 
The metamodels work by mimicking the output behaviour of the simulator as an analytical function of its inputs. While simulators often have complex internal relationships and dynamics, they can be treated as a ‘black-box' function with no clear mathematical formula. Nonetheless, the ‘emergent behaviour', resulting from the simulators' inner interactions and dynamics that evolve over time, is what the metamodels approximate. 

Metamodelling provides a framework to explore the output behaviour of the simulator efficiently, and this has been applied to many problems in transportation, \textit{inter alia}, for calibration of, e.g. origin-destination demand for large networks \citep{dantsuji2022novel}, day-to-day dynamics \citep{cheng2019surrogate}, and, more generally, a microscopic traffic simulator \citep{ciuffo2014sensitivity}, and for optimisation of, e.g. bus lane allocation \citep{li2022simulation} and traffic signal control \citep{osorio2013simulation}. However, in the scope of ATM research, this is quite recent~\citep{riis2021active,raquel2022rnest,cano2023nostromo}.

Given a metamodel, it is possible to evaluate thousands of simulation scenarios in a few seconds. Thus, metamodelling enables efficient exploration of the output behaviour of a simulator using computationally fast statistical models. However, by approximating the simulator with a machine learning model, we have, in practice, effectively replaced one black-box model with another, and thus the internal behaviour and its interactions are still not observable.\\

\textbf{Explainable metamodel}\quad Recently, \cite{riis2022explainablemetamodel} proposed an explainable metamodel by augmenting the metamodel with SHapley Additive exPlanations (SHAP) values \citep{lundberg2017unified}, enabling the extraction of significantly greater insights from the metamodel. The results show that the proposed methodology can effectively make simulators and their results more explainable, facilitating the interpretation of the obtained associated emergent behaviour and opening new opportunities for novel performance assessment processes in ATM. However, as \cite{riis2022explainablemetamodel} consider, a metamodel based on a fixed data set - a single design of experiment - consisting of 50,000 simulations, they only show the usage of SHAP values given a very large computational budget. In practice, the available computational budget is usually much smaller, and the real potential of metamodels is only really obtainable if the metamodel is constructed using active learning \citep{settles.tr09}, i.e. using active learning metamodels \citep{batchactivelearningAntunes,riis2022explainablemetamodel,raquel2022rnest}.\\

\textbf{Our contribution}\quad 
In this paper, we make a significant contribution to the field of applied metamodelling for air traffic management performance simulations. The main contribution lies in the introduction of our eXplainable Active Learning Metamodel (XALM), a three-step framework that integrates active learning, metamodelling, and SHAP values. XALM efficiently uncovers hidden relationships among input and output variables in complex simulated ATM systems, supporting ATM decision-making. We show the predictive performance of XALM through our experiments, achieving comparable results to the XGBoost metamodel while using fewer simulations. Furthermore, XALM exhibits superior explanatory capabilities compared to non-active learning metamodels. The practical application of XALM to a real-world scenario in Paris Charles de Gaulle airport, extending the arrival manager's range and scope through the analysis of six variables, highlights its effectiveness in enhancing simulation interpretability and understanding variable interactions. We also discuss two practical approaches to further reduce computational burden: the introduction of a stopping criterion for active learning based on metamodel uncertainty and the reuse of queried simulations for different key performance indicators.




\section{Background} 
In this section, we present the three distinct techniques and components, which we use in the proposed explainable active learning metamodel. We explain 1) the active learning approach, 2) the backbone model in our metamodel, namely, the Gaussian process, and lastly, 3) the explainable component in the form of the SHAP values. 

First, however, we present the notation used in the following sections. In the context of machine learning, a dataset is composed of individual data points, often organised in a matrix form, where the columns represent the dimensions (also known as features) and the rows the observations, additionally being either labelled or unlabelled. A labelled data point contains both an input feature vector, denoted by $\bm{x}$, and its corresponding output label, denoted by $y$, whereas an unlabelled data point only contains the features $\bm{x}$. In the case of data generated from a simulator, acquiring an unlabelled data point $\bm{x}$ is usually straightforward (specify the input variables to the simulator), whereas obtaining a labelled data point $(\bm{x},y)$ requires running the simulator with the input $\bm{x}$ to obtain the corresponding output $y$ (here the key performance indicator computed by the simulator).

\subsection{Active learning}
While a metamodel is computationally fast compared to the corresponding simulator, the metamodel can only approximate the simulator by actually `seeing' simulations from the simulator, meaning that to use the metamodel to avoid running the computationally expensive simulations, one first must run a few sets of simulations on which the metamodel will be trained. The framework of \textit{active learning} provides an efficient way to choose which simulations to run in order to make a good approximation in terms of high prediction performance while using the fewest resources spent on simulations.

Active learning aims to select the most informative data points sequentially, to improve model prediction performance and control costs. The learning process is guided by a label provider, also called the oracle (in our case, the simulator), which provides labelled data instances that are incorporated into the training data set. The oracle must be permanently accessible to be queried on-demand by the learning algorithm or model and should have the ability to generate labelled instances consistently from the ground truth function defining the process under study. Formally, and following the notation introduced by \cite{li2006confidence}, the five fundamental entities of an active learning system are the labelled training set $\mathcal{L}$, the set of unlabelled data points $\mathcal{U}$, the statistical model $\mathcal{M}$ (in our case, the metamodel), the oracle $\mathcal{O}$, and the query function $\mathcal{Q}$, which encodes the strategies and criteria for selecting informative instances from $\mathcal{U}$ to add to $\mathcal{L}$.

For a specific task, both the oracle, the labelled training set, and the unlabelled data points are often defined by the problem under study, whereas the querying strategy $\mathcal{Q}$ and the model $\mathcal{M}$ must be adapted for the task under study. 
Active learning uses different querying strategies (also known as acquisition functions) $\mathcal{Q}$, including uncertainty sampling \citep{MacKay1992}, query-by-committee \citep{RayChaudhuri1995}, 
and variance reduction \citep{gramacy2020surrogates}. 
These strategies rely either on the measure of informativeness or the identification of uncertainty regions to select the most informative instances to be added to the labelled training set.

\subsection{Gaussian processes}
Gaussian processes (GPs) are kernel methods that can be utilised for regression problems in a comparable way to linear regression. They are highly versatile and offer a Bayesian inference framework, being frequently used in both active learning and metamodelling applications. GPs have gained immense popularity for metamodelling tasks across various fields due to their modelling flexibility \citep{van2004kriging,kleijnen2009kriging,gramacy2020surrogates}. While initially used for deterministic simulations, the application of GPs was eventually widened to also include stochastic simulation settings \citep{van2003kriging}. Currently, despite vast developments in the machine learning field, traditional GPs, or variations thereof, stand as a common default choice for metamodelling~\citep{erickson2018comparison,yue2020active,knudde2020hierarchical,jiang2022gaussian,sauer2023non}.
For a comprehensive introduction to GPs, mostly from a machine learning viewpoint, refer to~\cite{Rasmussen2005}.

A GP is a stochastic function fully defined by a mean function $m(\cdot)$ and a covariance function (often called a kernel) $k(\cdot, \cdot)$. Given the data $ \mathcal{L}=(X, \bm{y})=\left\{\bm{x}_{i}, y_{i}\right\}_{i=1}^{N}$, where $y_i$ is the corrupted observations of some latent function values $\bm{f}$ with Gaussian noise $\varepsilon$, i.e.
$
    y_i = f_i + \varepsilon_i,~ \varepsilon_i \in \mathcal{N}(0, \sigma_\varepsilon^2),
$
a GP is typically denoted as $\mathcal{GP}(m_{\bm{f}}(\bm{x}), k_{\bm{f}}(\bm{x},\bm{x}'))$.
It is common practice to set the mean function $m_{\bm{f}}(\bm{x})$ equal to the zero-value vector 
and thus, the GP is fully determined by the kernel $k_{\bm{f}}(\bm{x},\bm{x}')$. For brevity, we will denote the kernel $K_{\boldsymbol\theta}$, which explicitly states that the kernel is parameterised with some hyperparameters $\boldsymbol\theta$.
Given $\boldsymbol\theta$, usually obtained via maximum likelihood estimation methods, the predictive posterior for unknown test inputs $X^\star$ is given by $p(\bm{f}^\star| \boldsymbol\theta, \bm{y}, X, X^\star) = \mathcal{N}(\boldsymbol\mu^\star, \boldsymbol\Sigma^\star)$ with
\begin{align}
    &\boldsymbol{\mu}^\star=K_{\boldsymbol\theta}^\star\left(K_{\boldsymbol\theta}+\sigma_{\varepsilon}^{2} \mathbb{I}\right)^{-1}{y}
    \quad\text{and}\quad
    \boldsymbol\Sigma^\star=K_{\boldsymbol\theta}^{\star \star}-K_{\boldsymbol\theta}^\star\left(K_{\boldsymbol\theta}+\sigma_{\varepsilon}^{2} \mathbb{I}\right)^{-1} K_{\boldsymbol\theta}^{\star \top}\, ,
\end{align}
where $K_{\boldsymbol\theta}^{\star \star}$ denotes the covariance matrix between the test inputs, and $K_{\boldsymbol\theta}^\star$ denotes the covariance matrix between the test inputs and training inputs.
For the kernel, we use the general applicable kernel, namely, the automatic relevance determination (ARD) radial-basis function (RBF) given by
\begin{align}
k\left(\bm{x}, \bm{x}'\right)=\sigma\exp \left(-||\bm{x}-\bm{x}'||^{2}/2 \boldsymbol{\ell}^{2}\right)\, ,
\end{align}
where $\boldsymbol{\ell}$ is a vector of length scales $\ell_1, ..., \ell_d$, one for each input dimension, and $\sigma$ is a scalar for the output variance.

\subsection{SHAP values}
SHapley Additive exPlanations (SHAP) is a unified framework for interpreting model predictions \citep{lundberg2017unified}. The SHAP values originate from game theory and have lately been used in machine learning to enhance model explainability. To explain a complex predictive model, SHAP constructs a simpler explanation model with $M$ simplified input variables. SHAP builds upon additive variables attribution methods, which have an explanation model that is a linear combination of binary variables, i.e. $g(\bm{x}) = \phi_0 + \sum_{i=1}^M \phi_i x_i$, where $x_i\in\{0,1\}^M$ and $\phi_i\in\mathbb{R}$. The method gives a unique solution and has two properties: 1)
    additivity of explanations, which means that the sum of all the feature attributions gives the output of the predictive model, allowing for the aggregation of feature contributions while maintaining consistency with the predictive model, even when dealing with categorical variables, and 2)
    consistency and local accuracy, which means that the output of the explanation model is consistent with the output of the predictive model, and if a feature has a positive impact on the prediction in the explanation model, it will not have a smaller impact in the predictive model.

There are various methods for enhancing model interpretability, including LIME and DeepLIFT. However, the SHAP values have emerged as a popular approach due to their unified framework, which aligns well with human intuition and computational efficiency \citep{lundberg2017unified}. This approach has been successfully used with tree ensemble methods like XGBoost through \texttt{TreeExplainers} \citep{lundberg2020local}, which utilise SHAP values to explain the entire model. Compared to previous techniques, \texttt{TreeExplainers} provides faster and more consistent results while capturing variable interactions, making it a valuable tool for interpreting tree-based ensemble methods \citep{lundberg2020local}.

For any non-tree-based method, the SHAP values are calculated using the \texttt{KernelExplainers}, which can explain any black-box model, including neural networks, support vector machines, and Gaussian processes \citep{lundberg2017unified}. This method uses a specially-weighted local linear regression to estimate the SHAP values and is slightly more computationally costly than the \texttt{TreeExplainers}.

\section{Methodology}
In this section, we introduce our explainable active learning metamodels and the ATM simulator under investigation. Then we describe the experimental setup, before presenting the details of the metamodel and the simulator.

\subsection{Explainable active learning metamodels}
\begin{figure}[t]
\centering
   \includegraphics[width=0.7\textwidth]{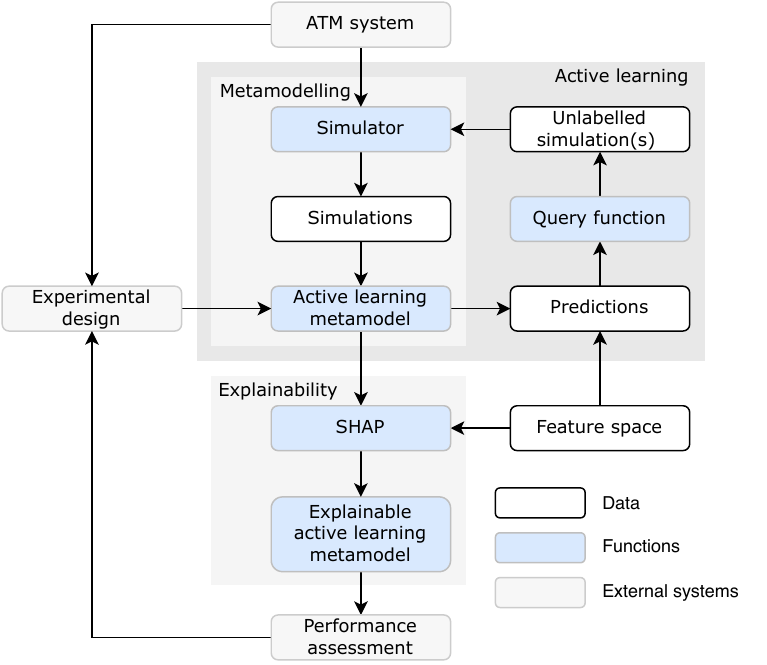}
  \caption{Explainable active learning metamodel}\label{fig:active_learning_metamodel}
\end{figure}

Our strategy to overcome the computational obstacles frequently encountered in simulation-based investigations consists of three main components. Firstly, we mitigate the computational complexity by applying a metamodelling framework that provides a fast approximation of the simulation outcomes and allows for predictions in unlabelled regions of the simulation input space. By predicting the output values for unlabelled input data points, a significant amount of simulation runs and time can be saved. Secondly, we adopt an active learning approach that focuses on selecting the most informative data points for posterior fitting in a more efficient way. Thirdly, after fitting the metamodel to a data set composed of pre-computed simulation results, we employ the SHAP method to enhance its interpretability, and hence, the transparency of the underlying simulator. We refer to this metamodel as an \emph{`explainable active learning metamodel'} since it gives an efficient model and offers a better understanding of the relationships between the input variables and KPIs, resulting in an increased level of comprehension regarding the problem under consideration. Our approach is illustrated in \autoref{fig:active_learning_metamodel}, and in the following, we describe the framework in more detail.

Active learning metamodels are metamodels that are fitted using the iterative process of active learning. As such, the methodology of active learning metamodels consists of two main stages, each consisting of two sequential steps as illustrated in \autoref{fig:active_learning_metamodel}. The first stage involves training the metamodel using the available \textit{simulations}  $\mathcal{L}$. Once the \textit{active learning metamodel} $\mathcal{M}$ is trained, it is used to predict outputs over the simulation \textit{feature space} $\mathcal{U}$. In the second stage, the query function $\mathcal{Q}$, uses the predictions for the feature space to select new unlabelled data points, denoted by $\bm{x}'$, that need to be evaluated by the \textit{simulator} $\mathcal{O}$. The simulator then provides new outputs, denoted by $y'$, corresponding to the selected data points from the previous step, which are then added to the current training set $\mathcal{L}$. The proposed methodology thus enables the efficient and effective construction of a metamodel using active learning. The aforementioned steps are executed repeatedly until a stopping criterion is met. This criterion can be defined based on the metamodel's performance, such as the reduction in error or improvement in accuracy, or by a predefined number of iterations based on the available resources, budget, and time.

As discussed earlier, active learning aims to improve model training efficiency and predictive performance by identifying the most informative data points while reducing data redundancy and computational resources. 
To create a fast and efficient approximation of the simulator, it is crucial that the underlying model of the metamodel can effectively model complex functions and operate well with small- and medium-sized data sets. Moreover, incorporating some uncertainty measures in the model can facilitate active learning, thereby improving the accuracy of the approximation. GPs are the preferred underlying model for creating metamodels due to their flexibility, ability to handle small data sets, and provision of uncertainty estimates, as well as their ability to model complex functions \citep{gramacy2020surrogates}. By utilising the GP, active learning can reduce data redundancy and computational resources while improving model training efficiency and predictive performance. For GPs, the most common query is the uncertainty sampling strategy \citep{MacKay1992}, which is, in such cases, equivalent to minimising the predictive variance or entropy.

Lastly, the trained active learning metamodel and the acquired set of labelled simulations are combined in SHAP to create the explainable active learning metamodel (XALM). Now, XALM can be used for the performance assessment by using it to predict and explain new scenarios by applying it to the unlabelled feature space.

\subsection{Simulator}
The metamodels presented in this article are trained on an existing simulator called `Mercury' \citep{delgado2019domino}, a passenger mobility model able to compute various metrics, based on European flights. The model, the indicators selected, the input variables and the experimental setup are described hereafter. In this paper, we focus on a case study that has been studied with Mercury in the past: the problem of Extended Arrival MANager (E-AMAN) for European airports \citep{SESARcatalogue}. In this case study, we extend the range (over the current maximum range) and scope of the standard AMAN for Paris Charles de Gaulle airport, a major hub in western Europe, and study the influence of this change on the flights.

In the following, we first describe the simulator, before describing in more detail the E-AMAN module used for the case study. We then describe the indicators computed by the simulator (those learned by the metamodel) and the input variables (the features of the metamodel).

\subsubsection{Core Mercury}

Mercury is a large-scale, agent-driven air mobility simulator. It has been developed over many years to simulate the movements of aircraft and passengers around Europe. The full description of the simulator can be found in \citet{Gurtner2021}, and its main characteristics are:
\begin{itemize}
    \item The model features several agents, including flights, Airline Operating Centres (AOC), the Network Manager, etc. Each agent has a private memory and private `functions'. 
    \item The simulation is based on events, e.g. flight departures, triggering responses from various agents (e.g. checking for missing passengers). Events are dynamically created, rescheduled, and destroyed by agents.
    \item Aircraft and passengers are tracked by the simulation, checking for turnaround processes, missing connections, etc.
    \item The simulation includes exogenous delays for flights (related to weather, for instance), and endogenous delays due to missed connections, missed curfews etc.
    \item The delays, and various other processes, such as aircraft turnaround, are stochastic: values are sampled from distributions (or historical data) chosen by the user.
    \item The simulator features the entire ECAC space, i.e. it simulates all flights departing or arriving in the ECAC area. Airport processes are simplified for extra-ECAC airports.
\end{itemize}


Mercury requires several sets of data to run, in particular flight plans, schedules and passenger itineraries. These data sets are described in \citet{Gurtner2021}. Note that only one day of schedules and passenger data are used\footnote{The data used are from 2014, as this is the latest, complete data set, containing flights, schedules and passenger data, that is available to us}. Hence, the different runs of the simulator can be seen as different instances of the same day of operations, randomising delays, turnaround times, regulations etc. 

\begin{figure}[!ht]
    \centering
    \includegraphics[width=0.65\textwidth]{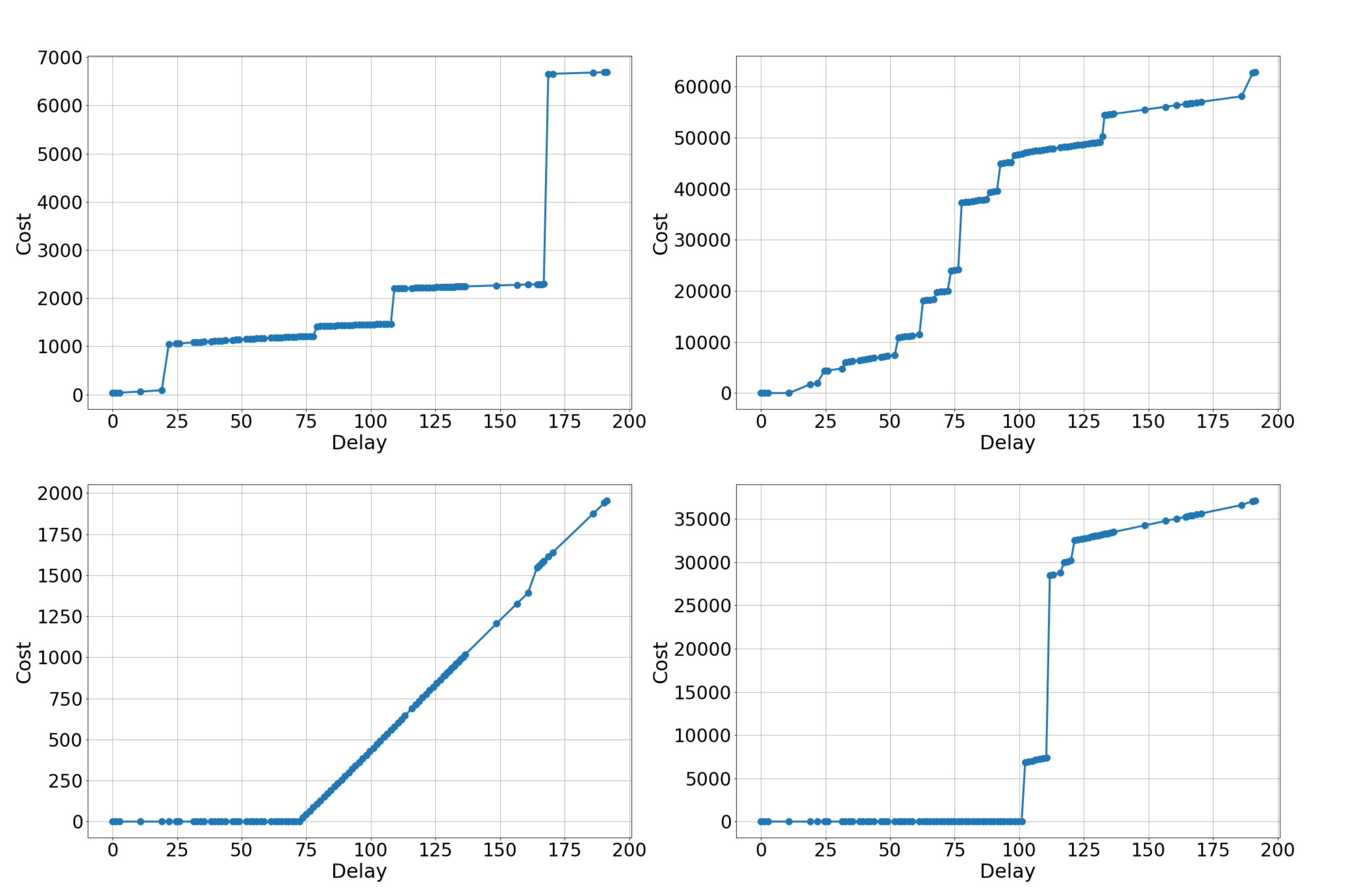}
    \caption{Examples of cost functions used in Mercury.}
    \label{fig:exmaples_cost_function}
\end{figure}

The agents within the simulator make decisions based on their own information and their inner objectives and logic. In particular, AOCs make various decisions related to flight management, including waiting for passengers or not, reallocating them, cancelling flights in some cases, etc. Their decisions are mostly driven by cost, and hence an adequate cost model is needed. Mercury includes an implementation of the costs model developed by \citet{Cook2011}, updated in \citet{Cook2015} and \citet{BEACON_D32}, which allow airlines to derive a cost function for each of their flights, i.e. a relationship between the delay incurred by a flight and the corresponding cost for the airline. Thanks to this model, the  cost function takes into account maintenance, crew (in an aggregated way), missed passenger connections (explicitly), turnaround delay (explicitly), and curfew costs (explicitly). As a result, very detailed delay cost functions for flights are obtained, such as those shown in Figure \ref{fig:exmaples_cost_function}.

For example, the bottom right graph shows two significant and several smaller jumps. The first significant jump is at $100$ minutes of delay and another one at about $110$ minutes. Such jumps usually indicate the presence of a group of passengers that would lose a connection in an onward flight and trigger compensation according to Regulation 261 \citep{eu2014compensation}. Another reason for a jump might be due to the need to change the crew when they exceeed their permissible hours, and another crew needs to be transferred to take the flight.

\subsubsection{E-AMAN module}

In this article, we selected a particular problem that has been studied in the past with Mercury: extending the range and scope of the Arrival Manager (AMAN).

The AMAN in Mercury is modelled as a stand-alone agent, distinct from the corresponding airport. Its default behaviour is fairly simple, ensuring that the runway capacity is not infringed, otherwise delaying flights en-route, or putting them in holding. The detection of the capacity infringement is done tactically every time a flight reaches a 100NM radius circle around the airport (the \emph{`tactical horizon'}, hereafter).

However, this process tends to cause unnecessary fuel burn, since flights could, in principle, absorb delay by slowing down much before the 100NM mark. Hence, the module implemented in Mercury features another horizon, called the \emph{planning horizon}, typically much larger than the tactical one, changing the modelled agent from AMAN to Extended-AMAN (E-AMAN). At the planning horizon, flights communicate their intent to the E-AMAN, which builds a `planning queue' of flights arriving at the airport. It then performs optimisation on this queue, minimising total delay. It then gives a specific command to the flight (either slow down, maintain speed, or speed up). We assume in the model that flights always follow this command, to the best of their abilities. We also assume that the flights are cleared regarding potential conflicts in the airspace that they will cross between their current position and arrival. We assume that flights departing within the planned horizon do not get any command, and just fly according to their flight plan, which is not known to the E-AMAN before they depart.

\subsubsection{Selection of input variables}
\label{sec:input_variables}

Mercury is highly parameterisable, from the exact forms of the probability distribution of delays to the price of fuel. In this case study, we are interested in testing the parameters linked to the E-AMAN itself, and also to macro-parameters that may show that the E-AMAN has different impacts when the system is in one state or another. For instance, we select the typical turnaround time as a parameter. Indeed, if the E-AMAN has a substantial effect, it should be even more substantial when the system is `tight', i.e. when the aircraft has a small amount of time between one flight and the other. 

The final parameters are presented in \autoref{tab:case_study}, as well as described below. Note that the parameters represent the inputs variables $x$ of the metamodels. In the table, we included the theoretical range of each parameter, the practical range, pushing some of them to the border of the realistic envelope, and the default value in the simulations when not specified otherwise. The parameters are the following:

\begin{itemize}
\item \textbf{Fuel price:} the approximated price of one kg of fuel. This has a direct impact on the airlines, since fuel represents a major share of their total costs. The fuel consumption itself varies in the simulations based on the holding and speed performed by the aircraft.

\item \textbf{Planning horizon:} the radius of the circle where flights enter the planned queue and the subsequent optimisation, as described above.

\item \textbf{Cruise uncertainty scale:} in the model, wind is modelled as a stochastic process modifying the ground speed of the aircraft (for a constant air speed). The wind has a central component (the mean) and associated variance. This parameter measures the scale of the variance compared to the baseline calibration. For instance, if this parameter is equal to 1.2, the standard deviation of speed is 20\% above its default value.

\item \textbf{Turnaround time scale:} in the model, an aircraft needs a certain time to be ready before its next flight. This time is a random number, drawn from a log-normal distribution. This parameter is the scale of the standard deviation of this distribution with respect to the baseline.

\item \textbf{Minimum connecting time scale:} passengers need a certain time to transfer from one aircraft to another if they are connecting. This mimimum connecting time is a fixed value for each airport, depending only on the type of connection made by the passenger (domestic, international, etc). This parameter represents the scale of this value with respect to the baseline.

\item \textbf{Claim rate:} this is the proportion of passengers that claim compensation according to Regulation 261. While this rate is currently quite high, due to increased public awareness and the growth of agencies supporting passenger claims, it was only around 14\% in 2014, the date of the dataset for which the model is running for this study.
\end{itemize}

\begin{table}[ht]
\caption{Input features used in the case studies.}
\label{tab:case_study}
\resizebox{1\textwidth}{!}{
\begin{tabular}{llllll}
\toprule
{\textbf{Feature}}                                                          & {\textbf{Short description}}                                                                                                  & {\textbf{Theoretical range}}                                                                                            & {\textbf{Practical range}} & {\textbf{Default}}     & \textbf{Unit}\\ \midrule
{Fuel price}                                                                 & {Price of one kg of fuel}                                                                                  & {{[}0, $\infty$)}                                                                                                          & {{[}0, 5{]}}               & {1}         & 2014 euros per kg          \\
{\begin{tabular}[c]{@{}l@{}}Planning\\ horizon\end{tabular}}                 & {\begin{tabular}[c]{@{}l@{}}Distance horizon where the E-AMAN\\  tries to optimise the arrival.\end{tabular}}      & {(100, $\infty$)}  & {{[}100, 1000{]}}          & 300 & Nautical miles\\
{\begin{tabular}[c]{@{}l@{}}Cruise\\ uncertainty\\ scale\end{tabular}}               & {\begin{tabular}[c]{@{}l@{}}Deviation in the aircraft speed\\  during cruise\end{tabular}}                              & {{[}0, $\infty$)}                                                                                                          & {{[}0, 10{]}}              & 1               & -     \\
{\begin{tabular}[c]{@{}l@{}}Turnaround\\ time scale\end{tabular}}           & {\begin{tabular}[c]{@{}l@{}}Scaler of mean of the distribution\\  of turnaround times\end{tabular}}                    & {{[}0, $\infty$)}                                                                                                          & {{[}0, 10{]}}              & {1}         & -         \\
{\begin{tabular}[c]{@{}l@{}}Minimum \\ connecting\\ time scale\end{tabular}} & {\begin{tabular}[c]{@{}l@{}}Scaler of mean of the distribution\\ of passenger minimum connecting\\  times\end{tabular}} & {{[}0, $\infty$)}                                                                                                          & {{[}0, 10{]}}              & {1}      & -              \\
{Claim rate}                                                                 & {\begin{tabular}[c]{@{}l@{}}Proportion of passengers claiming\\  compensation\end{tabular}}                             & {{[}0, 1{]}}                                                                                                            & {{[}0, 1{]}}               & {0.14} & - \\
\bottomrule
\end{tabular}
}
\end{table}

\subsubsection{Selection of key performance indicators (KPIs)}
\label{sec:PIs}
A characteristic of Mercury (and of any micro-simulator) is to be able to compute very low-level, disaggregated metrics. This profusion of numbers can be consolidated into a few indicators that are meant to estimate the efficiency of the system from different points of view. As a test case, we selected the following KPIs. Note, that theses KPI's represents the $y$ variables of the explainable active learning metamodels .

\begin{itemize}
\item \textbf{Flight arrival delay }(in minutes): computed by comparing the actual arrival time with the scheduled arrival time for each flight. The indicator is the average of this difference over all flights in the simulation.

\item \textbf{Flight departure delay} (in minutes): computed similarly to arrival delay, with departure times.

\item \textbf{Passenger arrival delay} (in minutes): computed by comparing the scheduled arrival of the last flight of each passenger, compared to the actual arrival. In case passengers cannot reach their final destination, a delay of 12 hours is considered.

\item \textbf{Planned absorbed delay} (in minutes): when the E-AMAN gives a `slow-down' or 'speed-up' command to a flight, the amount of time the flight would absorb en-route with the change of speed is computed. This is the time delay that aircraft could absorb while en-route, instead of being sent to the holding pattern before landing (holding consumes more fuel than cruise). This indicator averages this time over all issued commands.

\item \textbf{Holding time} (in minutes): the average of the time that the flights spend in holding.

\item \textbf{Fuel cost} (in 2014 euros): the average cost of the total fuel consumed by flights. Mercury computes this using the BADA3 \citep{nuic2010bada} model for fuel consumption and multiplying by the cost of the fuel.
\end{itemize}

\subsection{Experimental setup}
\subsubsection{Simulator}
Mercury is (here) run on 2014 data from Paris Charles de Gaulle airport, a major hub airport in Europe. All historical flights arriving to, and departing from, this airport are considered. Passengers on these flights may or may not connect at this airport, depending on the historical passenger itineraries. Exogenous delays are set to a high level in order to stress the system and better study the impact of the E-AMAN mechanism. 

This model is then considered as a black box for the metamodel, which can interact with it only by asking for a new simulation on a new set of inputs (only those described in \ref{sec:input_variables}) and getting in return the selected KPIs (only those described in \ref{sec:PIs}).

\subsubsection{Explainable active learning metamodels}
For our backbone model in the explainable active learning metamodel, we use a zero-mean GP with the general applicable ARD RBF kernel. In each iteration of the active learning loop, the inputs are rescaled to the unit cube $[0,1]^6$, and the outputs are standardised to have zero mean and unit variance. The initial data sets consist of ten data points chosen randomly, and in each iteration, one data point is queried. The unlabelled pool $\mathcal{U}$ consists of $50,000$ data points from the input space given by practical ranges in \autoref{tab:case_study}, and the test set consists of another $10,000$ simulations randomly sampled from the same space. In each active learning iteration, the GP is optimised for 300 gradient steps with Adam \citep{Kingma2015} with a learning rate of 0.1 and early stopping if the performance does not improve for 15 iterations. The GPs are implemented in GPyTorch \citep{Gardner2018}. As there exists no `ground' truth for the explainability of a model, we instead compute our own set of `ground' true values, using the SHAP values computed with XGBoost trained on 50,000 data points. The data set of the SHAP values consists of $1,000$ data points, and in each active learning iteration, we benchmark the explainability by comparing the SHAP values of the current metamodel with these values. For details on XGBoost and the optimisation thereof, see \autoref{app:xgboost}.

\subsection{Experiments on model performance}
In this section, we describe the three experiments in which we evaluate the performance of the explainable active learning metamodels. We make one metamodel per KPI, and all experiments are repeated 30 times with different initial data sets. 

\paragraph{Experiment 1: The Gaussian process as a metamodel} 
We test if the performance of the GP is on a par with the XGBoost trained on the extensive simulation study using 50,000 simulations \citep{riis2022explainablemetamodel}.
We evaluate the performance of a Gaussian process trained with 30, 100, and 1,000 simulations against two baselines: the lower baseline \textit{mean predictor}, which on new data points predicts the mean of the observed data points, and the upper baseline XGBoost trained on 50,000 data points, which represent an estimate of the best possible performance. Since the Gaussian process is computationally expensive to train due to the matrix inversion, we limit the largest number of data points used to fit the GP to 1,000 data points. Therefore, we also compare the performance against XGBoost fitted on 1,000 data points.

\paragraph{Experiment 2: Active learning vs. passive learning} 
We test if the active learning approach outperforms the random sampling, also known as passive learning. We report the performance of a Gaussian process trained  with active learning after querying a total of 30 and 100 simulations against the Gaussian process trained on a random sample of 30 and 100 simulations, respectively. Furthermore, to not only be dependent on those two specific values, we also visually compare the performance of active and passive learning by investigating the loss curves to show the effect of active learning for all active learning iterations up to a total dataset of 100 simulations.

\paragraph{Experiment 3: Active learning for explainable metamodels}
We test if the active learning approach is suitable when the objective is not to increase the performance of the model directly but to obtain SHAP values that correctly explain the predictions of the models. Similarly to experiment 2, we evaluate the performance, in terms of predictive accuracy for the SHAP values, after 30 and 100 simulations, comparing active and passive learning, as well as visual inspection of the loss curves.

Overall, we consider four different metrics to evaluate the performance of the models. We use the three metrics, the root mean square error (RMSE), relative root square error (RRSE), and the Pearson correlation, to evaluate the predictive performance. If we have the data set $\mathcal{L} = \{\bm{x}_i, y_i\}_i^N$ and the model $f$, such that $f(\bm{x}_i)=\hat{y}$ is the prediction of the model, then RMSE and RRSE is 
\begin{equation}
    \text{RMSE} = \sqrt{\dfrac{1}{N}\sum_i^N(\hat{y}_i-y_i)^2}
    \qquad\text{and}\qquad
    \text{RRSE} = \sqrt{\dfrac{\sum_i^N(\hat{y}_i-y_i)^2}{\sum_i^N(y_i-\bar{y})^2}}\quad\text{with}~\bar{y}=\dfrac{1}{N}\sum_i^Ny_i.
\end{equation}
The last metric is the computational time of the full pipeline, including the simulation and the training time of the models as well as the active learning.

\subsection{Analysis of the explainable active learning metamodels}
Our explainable active learning metamodels can provide multiple insights into what the model is predicting and, thus, how the outputs of the simulator behave: firstly, due to the explainable component of the SHAP values; and secondly, due to the fast inference of the metamodel, allowing for predicting the output of unseen simulation scenarios, accompanied by the inherent epistemic and aleatoric uncertainty estimates in the Gaussian process. We did an extensive pre-analysis (not included in the paper) to find any particular patterns and trends of interest, using the two aforementioned methods. However, in this paper, we only include a condensed analysis, which highlights the main findings. Nonetheless, we recommend the reader to have a look at the documentation of the SHAP values\footnote{Various visualisations of SHAP values: \url{https://shap-lrjball.readthedocs.io/en/latest/examples.html\#plots-examples}}.

In our analysis, we will use the SHAP values to summarise the feature contributions and highlight any particular impactful features, followed by individual plots, showing the trends of such features. These plots will be supported by plots showing the predictions of the metamodel with confidence and predictions interval (obtained through the epistemic and aleatoric uncertainty estimates from the GP), computed by fixing all features to their default value (cf. \autoref{tab:case_study}), and only changing the variable under investigation. Depending on the presence of any interactions between the features, we will include such by using different colours in the individual plots of the features as well as in the predictions of the metamodels. With the analysis, we not only aim to discover the underlying behaviour of the simulator Mercury, but also show how the explainable active learning metamodels can be used in a broader scope. Thus, we will also go into detail about how to interpret the explainable components of the metamodels. 

\section{Results}
We first benchmark the active learning with Gaussian processes by comparing the performance across a different number of simulations and against passive learning. 

\begin{table}[!ht]
    \footnotesize
    \centering
    \caption{Performance of the metamodels. The standard deviation is given in parentheses, computed across 30 repetitions.} 
    \label{tab:performance}
    \begin{tabular}{llrrrrl}
    \toprule
    \textbf{KPI} & \textbf{Model} & \textbf{\#sim} & \textbf{RMSE} & \textbf{RRSE} &\textbf{Correlation}\\
    \midrule
    {\textbf{Flight arrival delay}}  &
    Mean predictor        &50000&   106.0   &  1.00  & - \\
    & XGBoost         & 50000        &   3.91 (0.00)   &  0.04 (0.00) & 1.0 (0.00) \\
    & XGBoost         & 1000        &   4.27 (0.04)   &  0.04 (0.00) & 1.0 (0.00)\\
    & GP & 1000  &  3.97 (0.01) &  0.04 (0.00) &  1.0 (0.00) \\
    & GP & 100  &  4.41 (0.32) & 0.04 (0.00) &  1.0 (0.00) \\
    & GP & 30  &   5.77 (0.89) & 0.05 (0.01) &  1.0 (0.00) \\
    \midrule
    {\textbf{Flight departure delay}}  &
    Mean predictor        &50000&   106.4   &  1.00  & - \\
    & XGBoost         & 50000        &   3.86 (0.00)   &  0.04 (0.00)  & 1.0 (0.00) \\
    & XGBoost         & 1000        &   4.23 (0.04)   &  0.04 (0.00)  & 1.0 (0.00) \\
    & GP & 1000 & 3.90 (0.01) & 0.04 (0.00) & 1.0 (0.00) \\
    & GP & 100  &  4.31 (0.23) & 0.04 (0.0) &  1.0 (0.00) \\
    & GP & 30  &   5.58 (0.93) & 0.05 (0.01) &  1.0 (0.00) \\
    \midrule
    {\textbf{Passenger arrival delay}} & 
    Mean predictor        &50000& 202.3     &   1.00  & - \\
    & XGBoost         & 50000        &  43.0 (0.03)   &    0.21 (0.00)  & .98 (0.00) \\
    & XGBoost         & 1000        &  46.2 (0.46)   &    0.23 (0.00)  & .97 (0.00) \\
    & GP & 1000 & 43.7 (0.17) & 0.22 (0.00) & .98 (0.00) \\
    & GP & 100  &   46.9 (1.14) & 0.23 (0.01) &  0.97 (0.00) \\
    & GP & 30  &   61.8 (15.72) & 0.31 (0.08) &  0.95 (0.03) \\
    \midrule
    {\textbf{Planned absorbed delay}}  & 
    Mean predictor        &50000&   0.242   &  1.00  & - \\
    & XGBoost         & 50000        &   0.093 (0.000)   &  0.39 (0.00)  & .92 (0.00) \\
    & XGBoost         & 1000        &   0.098 (0.001)   &  0.41 (0.00)  & .92 (0.00) \\
    & GP & 1000 & 0.100 (0.001) & 0.41 (0.00) & 92 (0.00) \\
    & GP & 100  & 0.105 (0.005) & 0.43 (0.02) &  0.9 (0.01) \\
    & GP & 30  &  0.150 (0.032) & 0.62 (0.13) &  0.8 (0.09) \\
    \midrule
    {\textbf{Holding time}}  &      
    Mean predictor        &50000&   0.204   &  1.00  & - \\
    & XGBoost         & 50000        &   0.160 (0.000)   &  0.78 (0.00) & .62 (0.00) \\
    & XGBoost         & 1000        &   0.166 (0.002)   &  0.81 (0.01) & .59  (0.01) \\
    & GP & 1000 & 0.163 (0.001) & 0.80 (0.00) & .60 (0.00) \\
    & GP & 100  &  0.177 (0.008) & 0.87 (0.04) &  0.52 (0.06) \\
    & GP & 30  &   0.213 (0.023) & 1.04 (0.11) &  0.3 (0.18) \\
    \midrule
    {\textbf{Fuel cost}}  &      
    Mean predictor        &50000&   26401   &  1.00  & - \\
    & XGBoost         & 50000        &   354 \phantom{0}(0.4)   &  0.01 (0.00)  & 1.0 (0.00) \\
    & XGBoost         & 1000        &   474 (10.6)   &  0.02 (0.00)  & 1.0 (0.00) \\
    & GP & 1000 & 362 \phantom{0}(2.6) & 0.01 (0.00) & 1.0 (0.00)\\
    & GP & 100  &  398 (19.0) & 0.02 (0.0) &  1.0 (0.0) \\
    & GP & 30  &   531 (97.9) & 0.02 (0.0) &  1.0 (0.0) \\
    \bottomrule
    \end{tabular}
\end{table}
\begin{table}[!ht]
    \centering
    \caption{Computational time (in hours) for the different models. The time includes the simulation time (2.5 minutes per simulation) and training time. $^*$For fuel cost, the time was 8.51 and 
    3.05 hr for the GP with 100 and 30 simulations, resp.}
    \label{tab:timing}
    \begin{tabular}{lrr}
    \toprule
    \textbf{Model} & \textbf{\#sim} & \textbf{Time [hr]}\\
    \midrule
    Mean predictor        &50000& 2083.3 \\
    XGBoost         & 50000  & 2083.3 \\
    XGBoost & 1000 & 41.7   \\
    GP & 1000 & 41.8 \\
    GP & 100   & 4.9$^*$ \\ 
    GP & 30    &1.4$^*$ \\
    \bottomrule
    \end{tabular}
\end{table}
\paragraph{Experiment 1}
In \autoref{tab:performance}, the predictive performance is quantified with the RMSE, RRSE, and Pearson correlation for the GPs and the baselines. As expected, all the XGBoost and GP models outperform the mean predictor, and XGBoost trained on 50,000 simulations is, with regards to the predictive performance, the best-performing metamodel across the six KPIs and the three metrics. 

If we look at the performance of the three GP models, we see that the more simulations the better across all KPIs, and furthermore, if we compare the performance of the GP trained on 1,000 simulations to XGBoost trained on the same number of simulations, we see that the GP surpasses the XGBoost for five out of the six KPIs (XGBoost is slightly better for the planned absorbed delay). Overall, we see that the performance of the GP is on a par with XGBoost, and thus, in terms of predictive performance, it is reasonable to use the GP. 

The computational time for the different models arises mainly from the simulation time (2.5 min per simulation) and is, thus, approximately constant across the different KPIs. \autoref{tab:timing} shows that for the model with the 50,000 simulations, it takes 2083 hours, or 87 days, in CPU time, whereas the model with 30 simulations only takes 1.4 hours. As a consequence of the simulation time being the bottleneck, it is definitely beneficial to consider how many simulations are needed to achieve the desired performance. In the next experiment, we benchmark the active learning approach, which exactly deals with this trade-off in performance and computational time.
In conclusion, it is possible to get reasonable predictive performance with a much smaller set of simulations, significantly reducing the computational burden of performance assessment using simulators.

\begin{table}[!ht]
    \footnotesize
    \centering
    \caption{Performance of the Gaussian processes trained with active and passive learning. Best performance is in bold.} 
    \label{tab:al_performance}
    \resizebox{\textwidth}{!}{
    \begin{tabular}{llrrrrr}
    \toprule
    & & \multicolumn{2}{c}{\textbf{RMSE (predictions)}}&\multicolumn{2}{c}{\textbf{RMSE (SHAP)}}\\ \cmidrule(lr){3-4}  \cmidrule(lr){5-6} 
    \textbf{KPI} & \textbf{\#sim} & \textbf{Active learning} & \textbf{Passive learning}& \textbf{Active learning} & \textbf{Passive learning} \\ 
    \midrule
    {{Flight arrival delay}} 
     & 100  &  \textbf{4.22 (0.08)} &  4.41 (0.32) &\textbf{1.43 (0.40)} &  2.45 (1.55) \\ 
     & 30  &   \textbf{5.24 (0.29)} &  5.77 (0.89) &\textbf{1.87 (0.61)} &  3.55 (1.60) \\ 
    \midrule
    {{Flight departure delay}} 
     & 100  &  \textbf{4.13 (0.09)} &  4.31 (0.23) &\textbf{1.41 (0.37)} &  2.98 (1.63) \\
     & 30  &   \textbf{5.17 (0.41)} &  5.58 (0.93) &\textbf{1.79 (0.86)} &  4.26 (2.82) \\
    \midrule
    {{Passenger arrival delay}} 
     & 100  &  47.2 (1.37) &  \textbf{46.9 (1.14)} &\textbf{6.82 (1.18)} &  8.03 (2.58) \\ 
     & 30  &   \textbf{54.1 (6.01)} &  61.8 (15.72) &\textbf{9.64 (2.65)} &  14.50 (5.50) \\
    \midrule
    {{Planned absorbed delay}}  
    & 100  & \textbf{0.11 (0.00)} &  \textbf{0.11 (0.01)} &0.02 (0.00) &  \textbf{0.01 (0.00)} \\
     & 30  &  \textbf{0.12 (0.01)} &  0.15 (0.03) &\textbf{0.03 (0.01)} &  \textbf{0.03 (0.01)} \\ 
    \midrule
    {{Holding time}}        
    & 100  &  \textbf{0.17 (0.00)} &  0.18 (0.01) &\textbf{0.02 (0.00)} &  \textbf{0.02 (0.00) }\\ 
    & 30  &   \textbf{0.20 (0.02)} &  0.21 (0.02) &\textbf{0.03 (0.01)} &  0.04 (0.01) \\ 
    \midrule
    {{Fuel cost}}      
    & 100  &  \textbf{391 (14.3)} &  398 (19.4) &678 (166) &  \textbf{640 (352)} \\ 
    & 30  &   \textbf{468 (48.9)} &  531 (97.9) &\textbf{748 (379)} &  857 (473) \\ 
    \bottomrule
    \end{tabular}
    }
\end{table}
\begin{figure}[!ht]
    \centering
    \includegraphics[width=\textwidth]{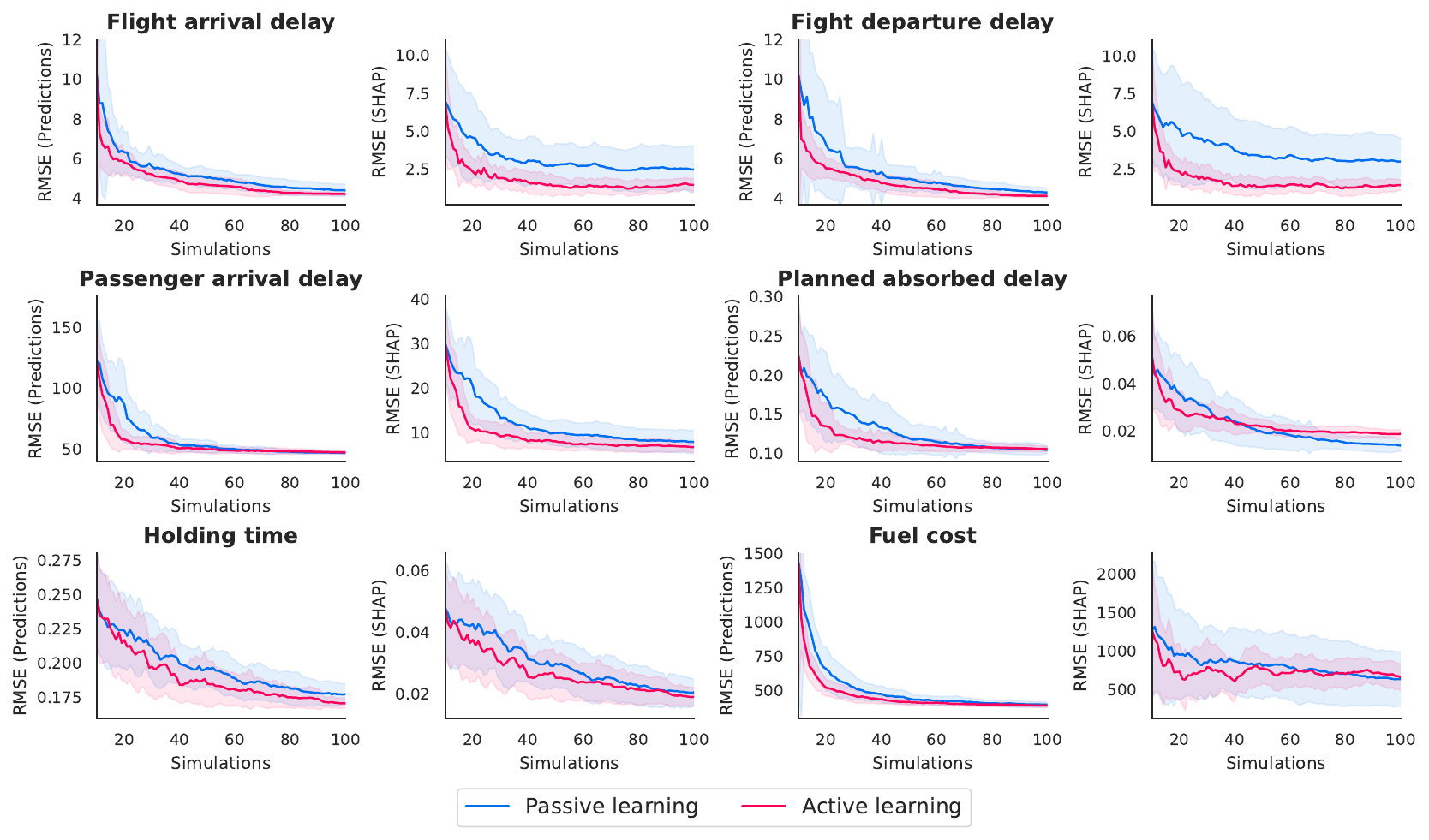}
    \caption{Active learning curves for the six KPIs. The figure shows the mean results of 30 repetitions, represented by lines, with the shaded area indicating the range of $\mu \pm \sigma$. For each KPI, the left panel shows the root mean square error (RMSE) of the predictions for the two acquisition functions over 90 acquisitions. The right panel displays the RMSE of the SHAP values compared to the SHAP values of the XGBoost using 50,000 simulations.}
    \label{fig:active_learning_curves}
\end{figure}
\paragraph{Experiment 2}
In \autoref{tab:al_performance}, the predictive performance is evaluated in terms of the RMSE for the GP trained with active learning and compared to the GP trained with passive learning (the results for passive learning are the same as in \autoref{tab:performance}). For the GPs trained on 30 simulations, the active learning approach outperforms passive learning across the KPIs. With 100 simulations, the difference in performance between the two approaches is smaller, and passive learning achieves the best performance for the two KPIs, passenger arrival delay and planned absorbed delay.

In \autoref{fig:active_learning_curves}, we have two plots for each KPI: one showing the active learning curves based on the RMSE for the predictions of the GPs, and one with the curves based on RMSE for SHAP values (the latter is discussed in experiment 3). The general trend across the KPIs is that the active learning curves are below the passive learning curves, showing how the former is more efficient than the latter. For all the KPIs, except holding time, we see that active learning is more efficient in the first iterations, where only a few simulations are used, compared to the late iterations, where the performance of active learning is only slightly better than passive learning, which is a natural pattern of active learning, well-known in the literature \citep{riis2021active,raquel2022rnest}. In summary, the active learning approach outperforms the passive learning approach.

\paragraph{Experiment 3}
In \autoref{tab:al_performance}, we see that the RMSE for the SHAP values obtained with GP trained on 30 simulations is best for active learning. When we use 100 simulations, active learning is better four out of six times. The active learning curves based on the RMSE for the SHAP values are shown in \autoref{fig:active_learning_curves} and exhibit similar patterns as for the active learning curves based on the RMSE for the predictions. In conclusion, active learning is indeed more efficient in sampling data points that give more precise SHAP values.

\subsection{Analysis}
In this section, we use the explainable active learning metamodels to describe the behaviour of the simulator, Mercury. First, we look at the overall feature contributions on the different KPIs, and then we investigate selected features' effects on single KPIs.
\mybox{Understanding the SHAP summary plots}{black!10}{white}{
The SHAP summary plots show the contribution of each feature on the corresponding output, i.e. KPI, based on the SHAP values. On the x-axis, we have the SHAP value, which is the impact on the model output, and the colours denote the value of each feature such that a blue and red colour corresponds to a low and high feature value, respectively. Note that the zero on the x-axis corresponds to the mean of the KPIs, and thus the SHAP values are computed with respect to this base value.
}
\begin{figure}[!ht]
    \centering
    \includegraphics[width=0.48\textwidth]{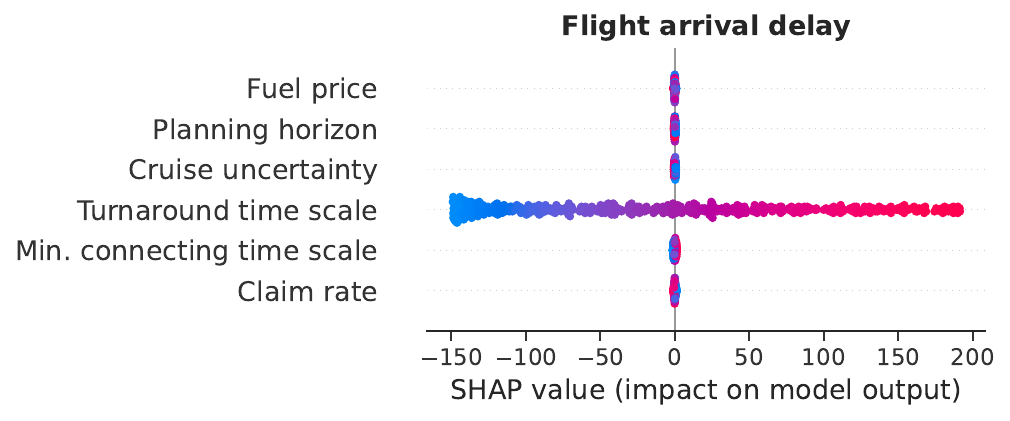}
    \includegraphics[width=0.48\textwidth]{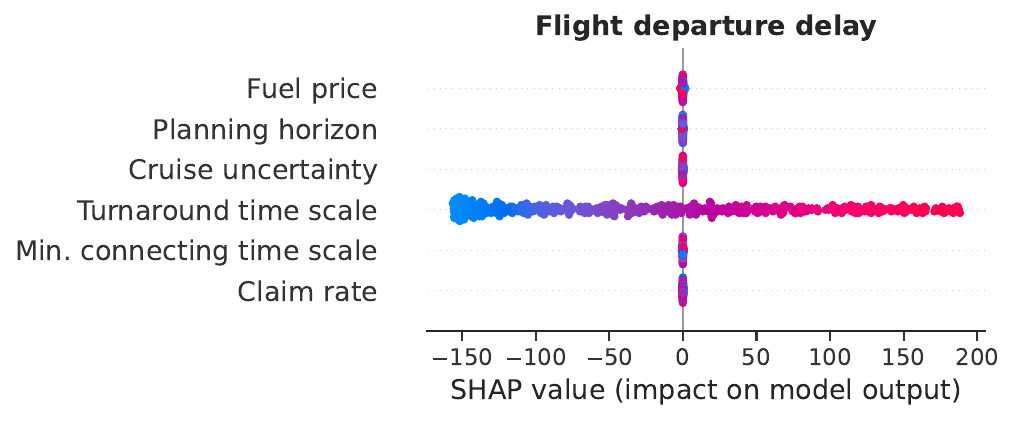}\\
    \includegraphics[width=0.48\textwidth]{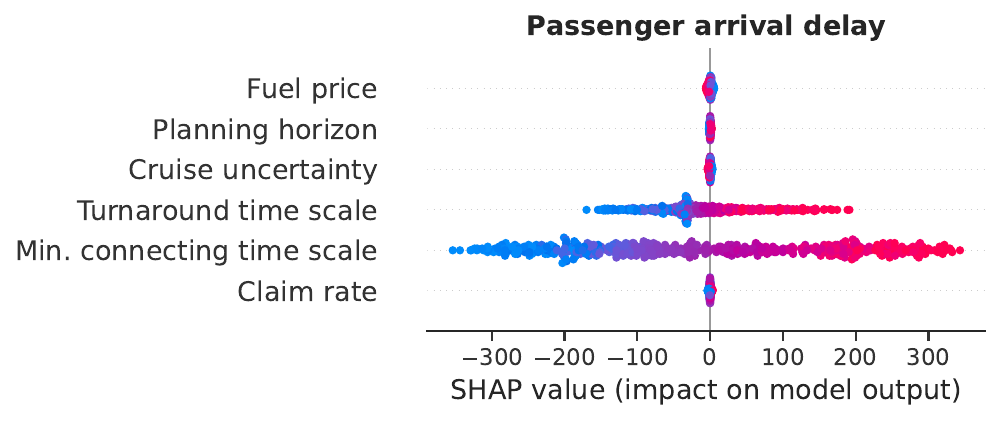}
    \includegraphics[width=0.48\textwidth]{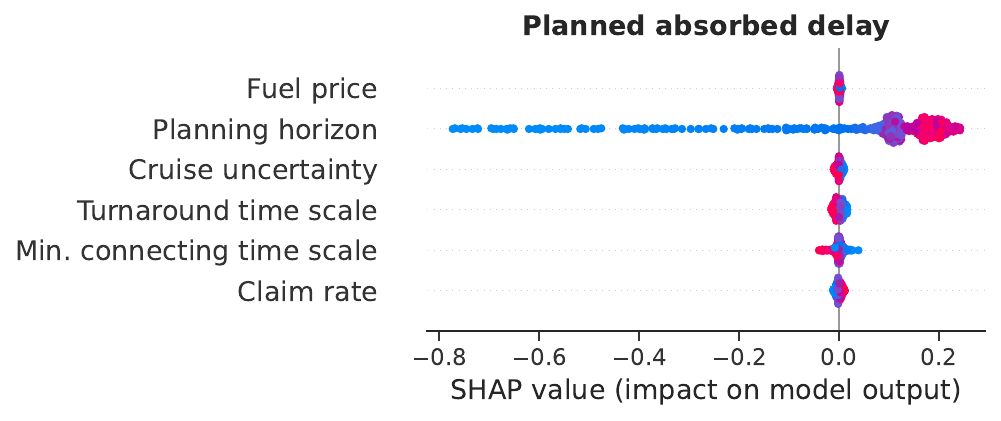}\\
    \includegraphics[width=0.48\textwidth]{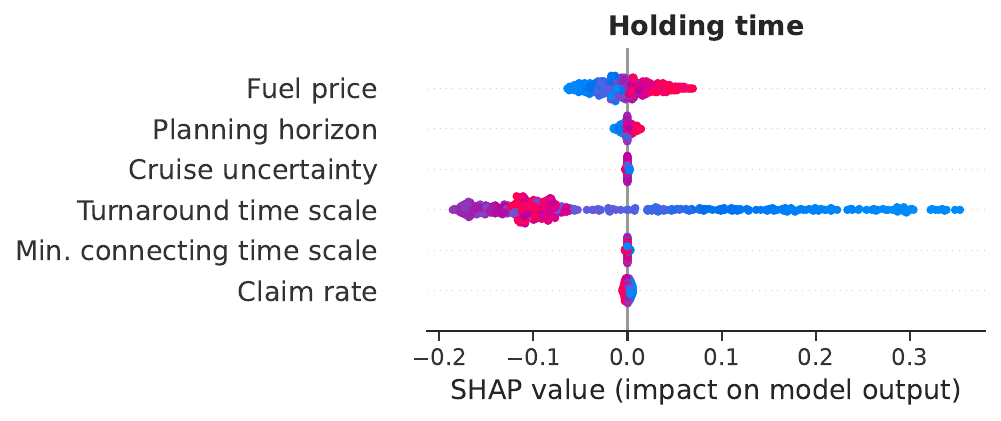}
    \includegraphics[width=0.48\textwidth]{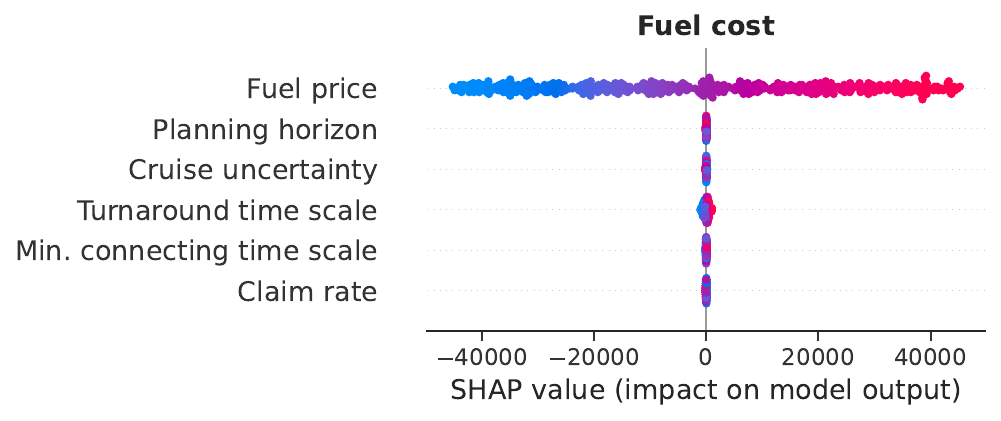}\\
    \includegraphics[width=.3\textwidth]{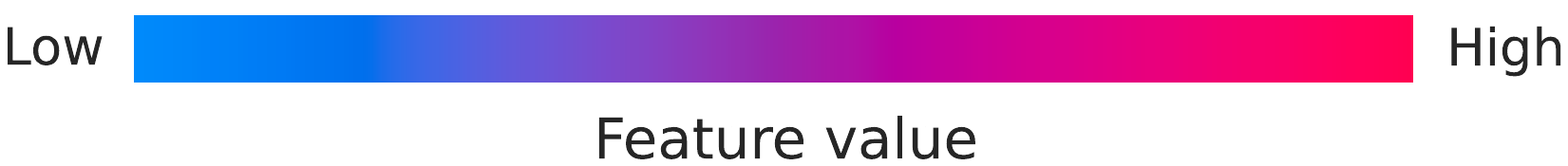}
    \caption{Feature contribution based on the SHAP values. The x-axis represents the impact on the model outputs, i.e. the KPIs. The colour represents the feature value of the features on the y-axis. All the plots are based on a GP fitted to 100 simulations.}
    \label{fig:shap_summary}
\end{figure}

\autoref{fig:shap_summary} shows the SHAP summary plots for the six KPIs. The flight arrival and departure delay are mainly influenced by the turnaround time, where a lower turnaround time gives a lower delay, and vice versa. The two KPIs, planned absorbed delay and fuel cost, are also mainly affected by single variables, namely, planning horizon and fuel price, respectively. The connection between the fuel price and fuel cost (see below) seems to be similar to that of the flight delays, whereas the planning horizon seems to have a particular effect on the planned absorbed delay when the planning horizon is low. Lastly, the passenger arrival delay and the holding time are affected by multiple features - patterns which we will discuss later. Overall, it should be noted that in this specific case study, the metamodel does not capture any effect of the cruise uncertainty or claim rate on any of the six KPIs under study.

In the following, we go through the contribution of specific features on the KPIs, starting with the simpler patterns arising from the fuel price's effect on the fuel cost and the planning horizon's effect on the planned absorbed delay. Afterwards, we investigate the more complicated cases with passenger arrival delay, and lastly, holding time. The plots for the effect of turnaround time on the flight arrival and departure delay are in \autoref{ap:tat_delays}.

\begin{figure}[!ht]
    \centering
    \includegraphics[width=0.48\textwidth]{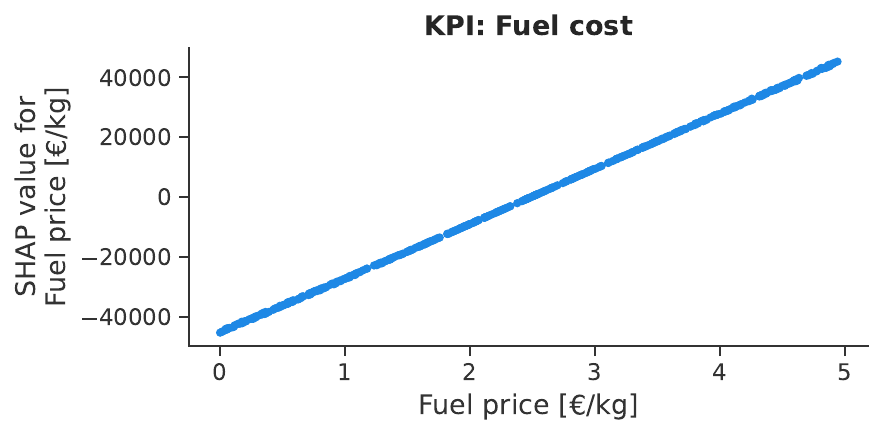}
    \includegraphics[width=0.48\textwidth]{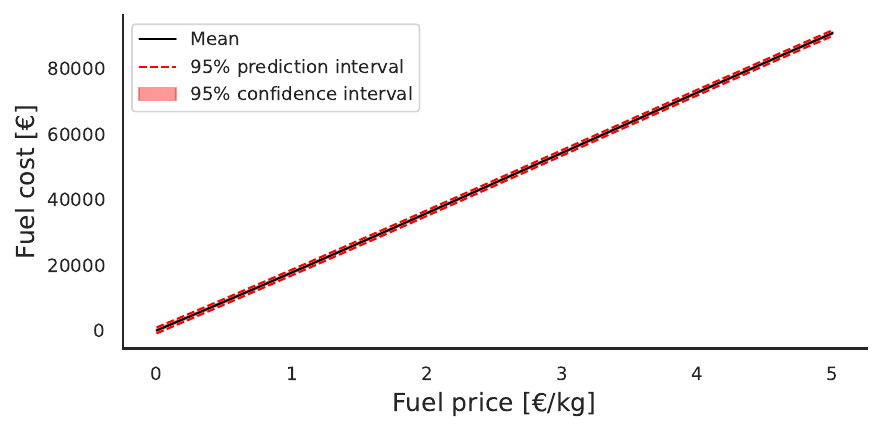}
    \caption{The impact of the fuel price on the fuel cost. \textbf{Left}: SHAP values. \textbf{Right}: Model predictions.}
    \label{fig:shapvalues_fuelcost}
\end{figure}

\paragraph{Fuel cost}
The left plot in \autoref{fig:shapvalues_fuelcost} shows the SHAP values fuel price and fuel cost. We observe a linear relationship between the input and the output, and for example that if the fuel price is near 0 \texteuro/kg, the fuel cost is the base value (the mean of the fuel cost) subtracted by \texteuro40,000. On the other hand, a fuel price close to 5 \texteuro/kg gives a fuel cost \texteuro40,000 above the base value. The SHAP values give a clear overview of the marginal effect of a feature on the output, but often it is also beneficial to look at the absolute effects for a specific set of features. The right plot in \autoref{fig:shapvalues_fuelcost} shows the prediction of the fuel cost when the features are set to the default values, and only the fuel price is changed. This shows clearly the overwhelming effect of the cost of fuel on the costs to the airlines, at least for the tactical costs incurred for these relatively low levels of delay (as refelected in \citet{Cook2015}, but also mindful that \textit{passenger} costs to the airline dominate at higher levels of delay). Note also that the upper bound of the fuel price modelled here is very high compared to today's.

\mybox{Understanding the predictions of the explainable active learning metamodel}{black!10}{white}{
With the GP metamodel, we can estimate the mean, the 95\% confidence interval, and the 95\% prediction interval. The mean prediction is the metamodel's estimate of the mean of the simulator's output if the simulator is run multiple times with the same features. The confidence interval shows how certain the model is about its mean prediction, and the prediction interval captures the stochasticity in the simulation output, such that if you run the simulator multiple times with the same features, 95\% of the simulations will be within the 95\% prediction interval.
}
%
Both the confidence and prediction interval coincide with the mean prediction, meaning that the metamodel is very certain about its mean predictions and that there is practically no stochasticity in the output.

\begin{figure}[!ht]
    \centering
    \includegraphics[width=0.48\textwidth]{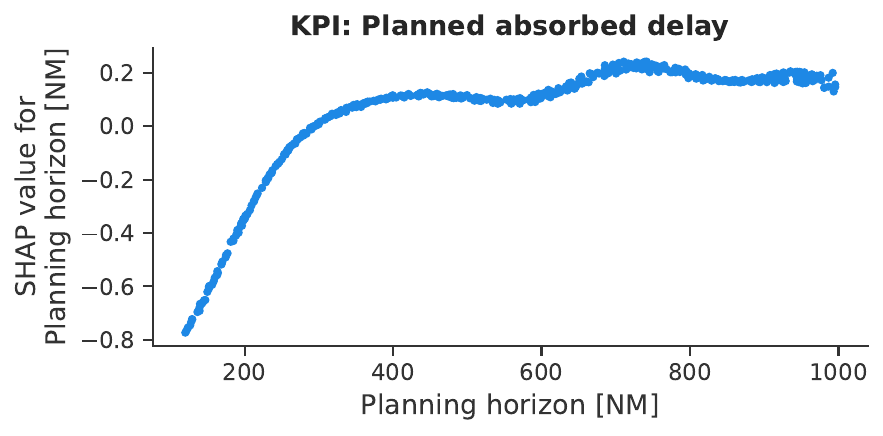}
    \includegraphics[width=0.48\textwidth]{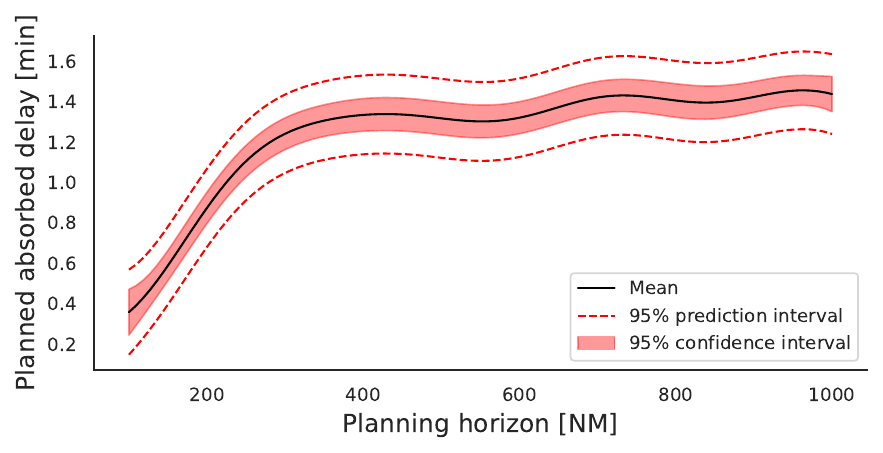}
    \caption{The impact of the planning horizon on the planned absorbed delay. \textbf{Left}: SHAP values. \textbf{Right}: Model predictions.}
    \label{fig:planned_absorbed_delay}
\end{figure}

\paragraph{Planned absorbed delay}
\autoref{fig:planned_absorbed_delay} shows how the marginal effect of the planning horizon (PH) has a linear relationship with the planned absorbed delay (PAD), when the PH goes from 100 to 300. For PH values higher than 300, the PAD seems to change periodically with a minor upward trend as the PH increases. However, before suggesting various hypotheses to explain this pattern, it is beneficial to examine the credibility of the metamodels through their predictions. In the right plot in \autoref{fig:planned_absorbed_delay}, we see the estimates of the mean, 95\% confidence interval, and 95\% prediction interval for the simulator's output with default features, only changing the planning horizon. The mean predictions have a similar pattern to that of the SHAP values, however, the 95\% confidence interval is rather wide and is, in fact, not excluding that the periodic pattern could be explained by a linear line. We also see that the prediction interval is even wider, meaning that the simulator has high stochasticity for the PAD.

Hence, PAD seems to follow a linear increase with respect to the PH, followed by a saturation (possibly slightly increasing up to 1000NM). This is an interesting pattern, since it shows that the theoretical efficiency of the E-AMAN increases at first (as more delay is planned to be absorbed) but reaches a maximum value after around 350NM. In other words, extending the E-AMAN horizon beyond this limit yields very little benefit.

\begin{figure}[!ht]
    \centering
    \includegraphics[width=0.48\textwidth]{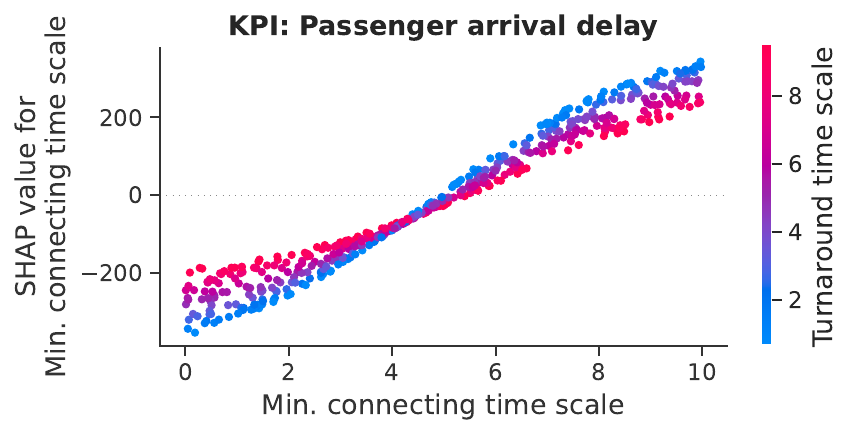}
    \includegraphics[width=0.48\textwidth]{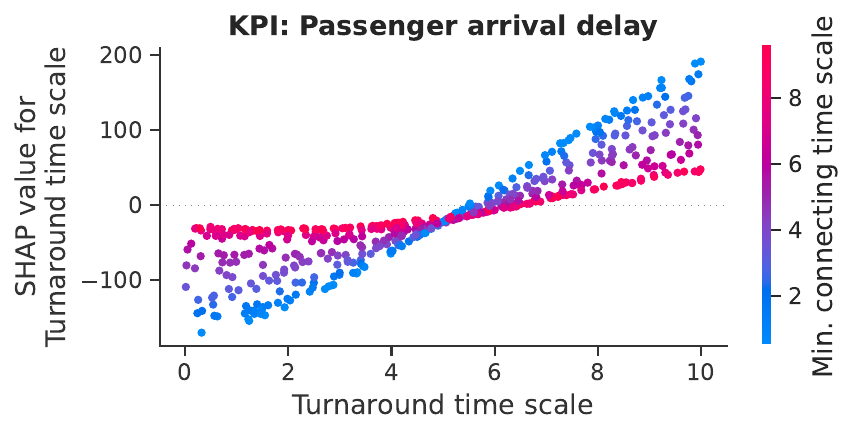}
    \caption{The two features with the highest impact on the passenger arrival delay.}
    \label{fig:passenger_arrival_delay}
\end{figure}

\paragraph{Passenger arrival delay}
In \autoref{fig:passenger_arrival_delay}, we see the two most important features for predicting the passenger arrival delay, namely, the minimum connecting time scale (MCT) and the turnaround time scale (TT). In the left plot, we have the MCT on the $x$-axis and the corresponding SHAP values on the $y$-axis. The colours denote the value of the feature TT in the same way as the colours in the SHAP summary plots, such that a blue point and red point correspond to a low and high TT, respectively. Irrespective of the colours, we see that the higher MCT, the higher the SHAP value, meaning that the higher the MCT, the more it contributes to a higher passenger arrival delay. The colours indicate the value of TT, showing that if TT is low, there is a more considerable change in the passenger arrival delay when MCT is increased compared to when TT is high. In other words, if TT is high, the value of MCT has a smaller impact on the passenger arrival delay. In the right plot, we investigate the same interaction, now having interchanged the MCT and TT. In general, the passenger arrival delay is linearly dependent on the TT. When MCT is low, the linear increase in the passenger arrival delay is higher with respect to TT, whereas, when MCT is high, the value of TT affects the passenger arrival delay less. 

The monotonic nature of the passenger delay against both MCT and TT is natural. Since the schedules are fixed in our experiments (and in general, at least during a season), increasing either the time the aircraft take to be prepared for the next rotation, or the time needed for passengers to connect to the subsequent flight, is going to result in more delays. In the first case, the next rotation is delayed, which may delay next rotations, and/or imply that passengers are going to miss their next connection. Conversely, increasing the TT will imply that more passengers are going to miss their flights when connecting. Both of these will increase the delay at passengers' final destination.

It was also expected that higher TT values decrease the impact of MCT on the delay, and vice versa. If, for example, aircraft take more time to be prepared for their next rotation, then passengers will have more time to reach them, \textit{ceteris paribus}.

In the following, we examine the two features' role in predicting the passenger arrival delay as well as their interactions by looking at the predictions from the metamodel. We incorporate the interaction into the plots by choosing three different values for the secondary feature such that we see, for example, how MCT affects the passenger arrival delay for three different values of the TT.

\begin{figure}[!ht]
    \centering
    \includegraphics[width=.95\textwidth]{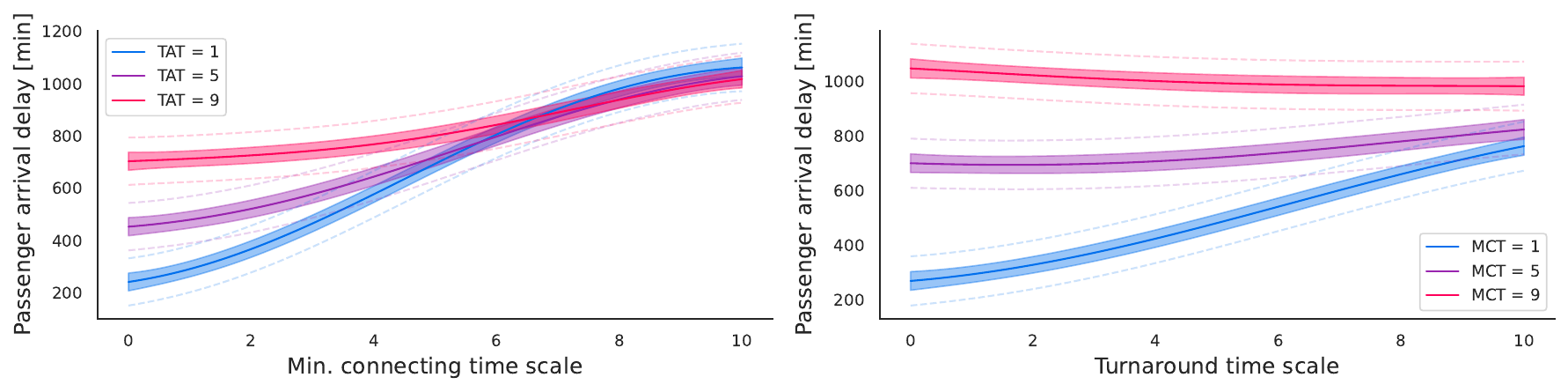}
    \caption{The two features with the highest impact on the passenger arrival delay.}
    \label{fig:passenger_arrival_delay_preds}
\end{figure}

The left plot in \autoref{fig:passenger_arrival_delay_preds} shows that the passenger arrival delay increases when MCT increases. When TT is low (blue line), the passenger arrival delay increases from 200 to 1000 as MCT is increased from 0 to 10. If TT is high (red line), the passenger arrival delay is already high for low values of MCT, and for MCT higher than six, the passenger arrival delay is only slightly affected by TT.
The right plot in \autoref{fig:passenger_arrival_delay_preds} shows that for a high MCT, the passenger arrival delay is around 1000 and is slightly decreasing as TT increases. Conversely, when MCT is low, TT has a high effect on the passenger arrival delay. In both plots, the confidence and prediction intervals are stable across the different values of MCT and TT, and we see that there is some stochasticity in the output of the passenger arrival delay. Note how the decrease in passenger arrival delay might seem contrary to the fact that the SHAP value's impact on the passenger arrival delay for TT is increasing as TT increases in \autoref{fig:passenger_arrival_delay}. However, the previously mentioned interaction with the minimum connecting time scale is causing this decrease. Consider the SHAP values in \autoref{fig:passenger_arrival_delay}: in the right plot, the red points represent the SHAP values for TT, when MCT is high, and ranges from -20 to 40, approximately linearly increasing with TT. Now, if we consider the left plot and examine the SHAP values for high values of MCT, e.g. MCT equal to 8, the SHAP values are ranging from approximately 180 to 350, as TT decreases. Thus, combining the SHAP values from TT and MCT yields a decrease in the passenger arrival delay as TT increases, as seen in the predictions in \autoref{fig:passenger_arrival_delay_preds}

\begin{figure}[!ht]
    \centering
    \includegraphics[width=0.48\textwidth]{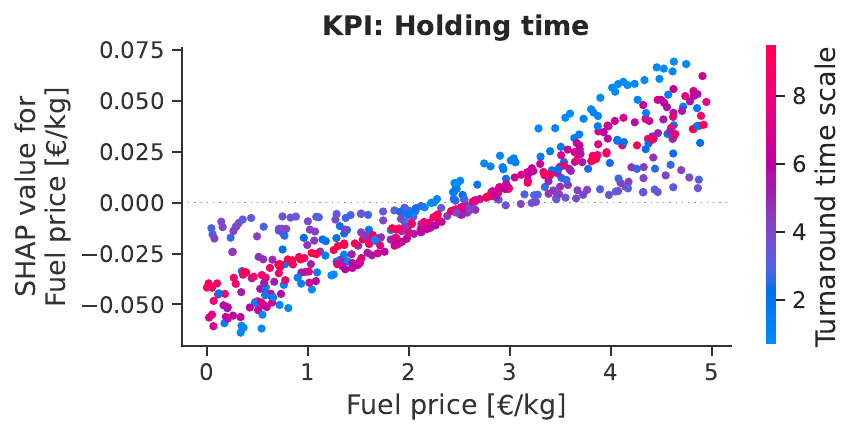}
    \includegraphics[width=0.48\textwidth]{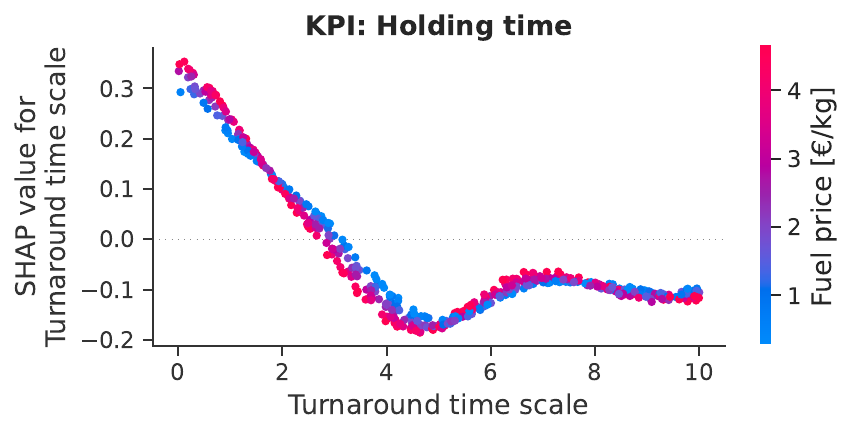}\\
    \caption{The two features with the highest impact on the holding time.}
    \label{fig:holdingtime}
\end{figure}

\paragraph{Holding time}
In \autoref{fig:holdingtime}, we see the two most important features for predicting the holding time, namely, the fuel price (FP) and the turnaround time scale (TT). For the SHAP values for the FP, we see that the higher the FP, the higher the holding time, although when the TT is around 5, we see only a small increase in the holding time. The fuel price has the highest impact on the holding time when TT is low. In the right plot, we see that the marginal effect of TT on the holding time has a negative linear relationship when TT is below 4. For TT higher than 4, we note an increase in the holding time before a minor decrease. Again, it is helpful to consider the validity of the SHAP values by looking at the credibility of the metamodel through its predictions of confidence intervals. 

\begin{figure}[!ht]
    \centering
    \includegraphics[width=0.9\textwidth]{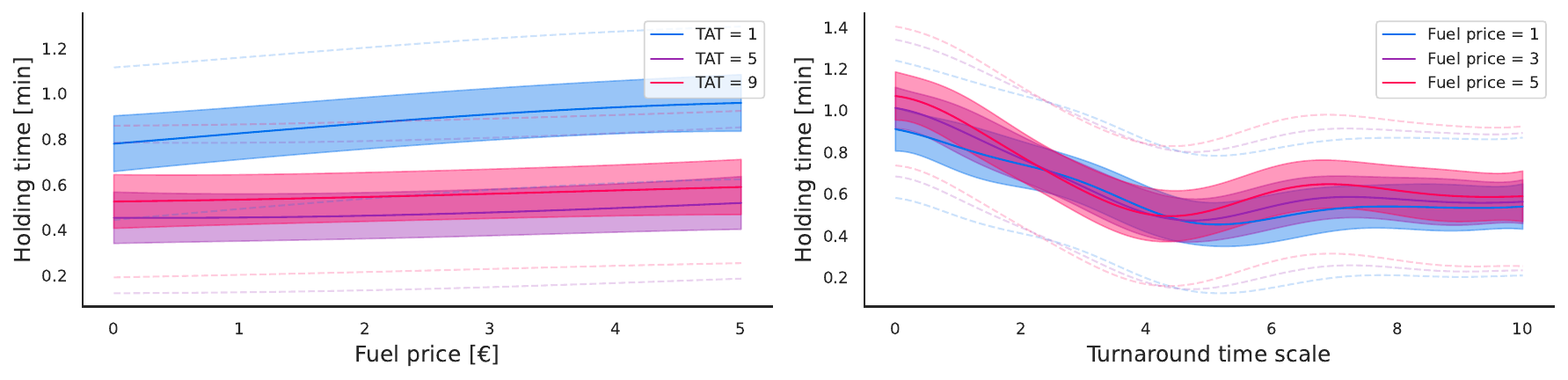}
    \caption{The two features with the highest impact on the passenger arrival delay.}
    \label{fig:holdingtime_preds}
\end{figure}

In the left plot in \autoref{fig:holdingtime_preds}, we see that the predictions also reflect the linear relationship between the fuel price and the holding time. However, we also note that both the confidence and prediction intervals are rather wide, where the former means that the metamodel is uncertain regarding the intercept and coefficient of the linear relationships, and the latter that there is high stochasticity in the holding time. In the right plot, we see a similar pattern in the confidence and prediction intervals, showing high uncertainty about the mean predictions and in particular the pattern in holding time for TT higher than 4.

Hence, the holding time is an increasing function of the fuel price, although less strongly related when the TT is higher. It is somewhat of a puzzling result since it should be expected to have less holding time when the fuel is more expensive. This is probably because the E-AMAN does not work as intended. Indeed, when the price of fuel is higher, the E-AMAN will try to give `slowing down' commands more often, to save fuel during cruise. However, it may be that the randomness of the trajectory means that the flight, when arriving at the tactical horizon, has lost the slot it was supposed to take when the command was issued. Hence, the flight then has to take extra holding time. 

This effect is slightly mitigated by the turnaround time, up to a certain level, because higher TTs mean less slack for the next flight. Hence, slowing down commands are less likely to be issued, decreasing the side effect of having to hold longer afterwards.

\section{Discussion}
\subsection{Practical considerations on stopping criterion for the active learning}
Often there is a computational budget that limits the analysis, making a stopping criterion based on the performance unnecessary. However, in some cases, it is beneficial to track the performance of the model in each iteration. In the machine learning community, it is common practice to evaluate the performance of a model on a hold-out test set. However, since active learning is all about reducing the amount of data points needed, we should rather avoid spending resources on creating such a data set. A common tool to utilise in such situations is cross-validation, which could be used to estimate the generalisation performance of the model, although it has a computational overhead. However, since the GPs have an intrinsic measure of the epistemic uncertainty in the form of the noise-free predictive uncertainty, a computational-free measure is to evaluate this and define the stopping criterion based on the epistemic uncertainty, or changes therein, to be below a certain threshold \cite{batchactivelearningAntunes,hino2020active}.

\begin{figure}[!t]
    \centering
    \includegraphics[width=0.7\textwidth]{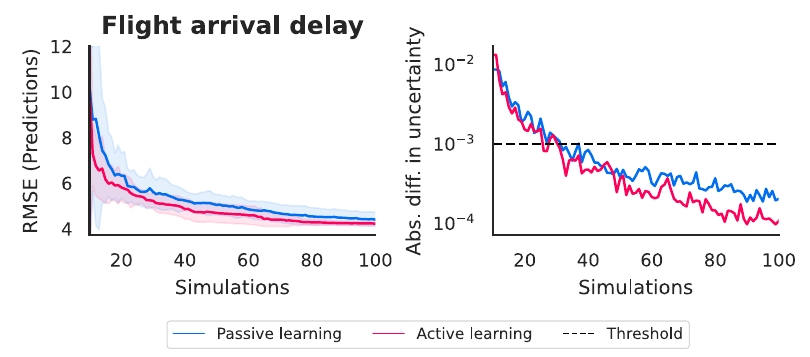}
    \caption{Active learning curves for the flight arrival delay. The figure shows the mean results of 30 repetitions, represented by lines, with the shaded area indicating the range of $\mu \pm \sigma$. \textbf{Left}: The root mean square error (RMSE) of the predictions for the two acquisition functions over 90 acquisitions. \textbf{Right}: The absolute difference in the epistemic uncertainty between the active learning iterations, where the epistemic
uncertainty for the model is predicted for the unlabelled data set.}
    \label{fig:minimize_variance_arrdel}
\end{figure}

In \autoref{fig:minimize_variance_arrdel}, we show the absolute difference in the epistemic uncertainty between the active learning iterations, where the epistemic uncertainty for the model is predicted for the unlabelled data set. Comparing it to the left plot in \autoref{fig:minimize_variance_arrdel}, we see that curves for the absolute difference in the uncertainty approximately follow that of the RMSE of the SHAP values. The curves for the other KPIs are in \autoref{app:minimize_variance}.

For a quantitative evaluation of how well the epistemic uncertainty estimates represent the RMSE of the predictions, we show the correlation coefficients $\rho$ between the two in \autoref{tab:correlation_betwee_epistemic_and_rmse}. The coefficients $\rho$ range from 0.69 to 0.93, and we thus see that the epistemic uncertainty from the GP is sufficiently correlated with the RMSE to be used as a proxy for the predictive performance.
\begin{table}[!t]
    \centering
    \caption{The mean (of 30 runs) correlation coefficients between the normalised noise-free predictive uncertainty and the RMSE of the predictions and RMSE of the SHAP values, computed across the 100 active learning iterations.}
    \label{tab:correlation_betwee_epistemic_and_rmse}
    \begin{tabular}{lrr}
    \toprule
    \textbf{KPI} &  $\rho_\text{RMSE~(pred)}$ & $\rho_\text{RMSE~(SHAP)}$ \\ 
    \midrule
    {{Flight arrival delay}} & 0.81 (0.10) &  0.68 (0.15) \\ 
    
    {{Flight departure delay}}& 0.82 (0.09) &  0.67 (0.17) \\ 
    
    {{Passenger arrival delay}}& 0.75 (0.16) &  0.74 (0.16) \\ 
    
    {{Planned absorbed delay}}  & 0.69 (0.31) &  0.57 (0.40) \\ 
    
    {{Holding time}} & 0.71 (0.20) &  0.69 (0.22) \\ 
    
    {{Fuel cost}} & 0.93 (0.04) &  0.25 (0.35) \\
    \bottomrule
    \end{tabular}
\end{table}

We now use the epistemic uncertainty estimate as a stopping criterion in practice. We define the stopping criterion, such that we either stop if a metric is below a certain value or after we have queried 100 simulations. We use the metric computed as the absolute difference between the normalised epistemic uncertainty, such that it is independent of the scale of the output. Additionally, to ensure convergence, we only stop if the value is below the threshold for three iterations in a row, and we set the threshold to 0.001. In \autoref{tab:stopping_criteria_results}, we see that the stopping criterion requires a different number of simulations for the different KPIs, where the flight arrival and departure delay together with fuel cost requires fewer than 17 simulations to achieve reasonable performance, whereas the holding time almost needs the full computational budget, on average, with 97 simulations.

\begin{table}[!ht]
    \centering
    \caption{The performance of the XALM using the stopping criterion, averaged across 30 experiments.}
    \label{tab:stopping_criteria_results}
    \begin{tabular}{lrrrr}
    \toprule
    \textbf{KPI} &  Simulations & RMSE (pred) & RMSE (SHAP) & Time [hr] \\ 
    \midrule
    {{Flight arrival delay}} & 16.3 (2.7) &  6.32 (1.10) &  1.88 (0.76) &  0.78 (0.13) \\ 
    
    {{Flight departure delay}}& 15.5 (1.5) &  6.27 (1.41) &  1.90 (0.91) &  0.73 (0.07) \\
    
    {{Passenger arrival delay}}& 26.5 (14.8) &  53.98 (5.89) &  9.26 (2.49) &  1.28 (0.75) \\ 
    
    {{Planned absorbed delay}}  & 41.5 (32.2) &  0.12 (0.01) &  0.03 (0.01) &  2.05 (1.62) \\ 
    
    {{Holding time}} & 96.4 (13.8) &  0.17 (0.01) &  0.03 (0.01) &  4.49 (0.66) \\ 
    
    {{Fuel cost}} & 13.1 (1.2) &  785 (193) &  795 (385) &  0.67 (0.13) \\ 
    \bottomrule
    \end{tabular}
\end{table}

\subsection{Reusing simulations across KPIs}
Often, a simulator outputs multiple KPIs at once, and thus for each simulation input variable vector $\bm{x}$, there are multiple outputs $\bm{y}$. This means that we can consider multiple KPIs simultaneously and create a single explainable active learning metamodel to predict all the KPIs of interest, and thus save a significant amount of the computational burden of the simulator. \cite{riis2021active} apply active learning such that new data points are acquired based on all the KPIs, resulting in new data points that add as much information about the KPIs as possible in each iteration. However, if we know \textit{a priori} that an output is more complex to learn and requires more simulations than the others, e.g. such as `holding time', in this specific case, we can instead perform active learning with respect to this KPI and then reuse the simulations to fit the metamodels for the other KPIs. 
\begin{table}[!ht]
    \centering
    \caption{The performance of the metamodels trained on the simulations queried by the XALM used for the `holding time'.}
    \label{tab:multiout_perf}
    \begin{tabular}{lrr}
    \toprule
    \textbf{KPI} &  RMSE~(pred) & RMSE~(SHAP) \\ 
    \midrule
    {{Flight arrival delay}} & 4.39 (0.38) &  1.72 (0.82) \\  
    
    {{Flight departure delay}}& 4.26 (0.35) &  1.69 (0.81) \\  
    
    {{Passenger arrival delay}}& 47.16 (2.33) &  7.70 (1.82) \\ 
    
    {{Planned absorbed delay}}  & 0.11 (0.01) &  0.03 (0.01) \\
    
    {{Holding time}} & 0.17 (0.01) &  0.03 (0.01) \\
    
    {{Fuel cost}} & 399 (26) &  815 (236) \\ 
    \bottomrule
    \end{tabular}
\end{table}
\autoref{tab:multiout_perf} shows the performance of the metamodels trained on the data points queried by the XALM used for the holding time. If we compare the numbers in \autoref{tab:multiout_perf} with the numbers in \autoref{tab:al_performance}, we see that the RMSEs for the predictions are slightly worse than for XALMs optimised for the specific KPIs, though still on a par with the passive sampling. If we compare the RMSEs for the SHAP values, our XALM optimised for the holding time is better than random sampling in three out of six KPIs and on a par for the other three. Reusing the simulations to train the metamodels for the other KPIs reduces the overall computational time from 10 to 4.5 hrs in total.


\section{Conclusions}
In this paper, we proposed and demonstrated XALM (eXplainable Active Learning Metamodel), a three-step framework that integrates active learning and SHAP values with simulation metamodels, thus addressing computational limitations while improving interpretability. XALM offers a unified modelling approach to efficiently discover the intricate hidden relationships between the input and output variables of a simulator. Through two experiments, based on a real-world ATM scenario (extended arrival management), we demonstrated the predictive performance of XALM, showing that it achieves comparable results to an XGBoost metamodel with significantly fewer simulations. 

In a third experiment on explainability, we highlighted the superior explanatory capabilities of XALM compared to non-active learning metamodels. We also discussed two practical approaches to further reduce computational burden by introducing a stopping criterion based on metamodel uncertainty and by reusing previously queried simulations for different key performance indicators.

Applying XALM to the ATM scenario using the Mercury simulator, we also explored the effects of extending the range and scope of the arrival manager by analysing six variables. The case study highlighted the effectiveness of XALM in enhancing the interpretability of simulations, enabling a deeper understanding of variable interactions. Our approach complements traditional simulation-based analyses, providing a practical solution to computational challenges and improving explainability.

The main metric associated with the cost of delay to the airlines is arrival delay (\citet{Cook2015}), which is itself, in Europe,  basically driven by departure delay (\citep{PRR_2022}), since there is relatively little en-route recovery in the European context. We discussed in our analyses (e.g. with reference to \autoref{fig:shap_summary}) how both flight arrival and departure delay are mainly influenced by the turnaround time, where a lower turnaround time gives a lower delay, and vice versa. This suggests the importance of both reactionary delay (e.g. whereby inbound aircraft delay causes an outbound delay) and airline-related delay (usually delay at the gate) in the network. We take a brief look at empirical data relating to delays to emphasise the relative importance of these factors in actual operations.

\begin{figure}[t]
\centering
   \includegraphics[width=0.7\textwidth]{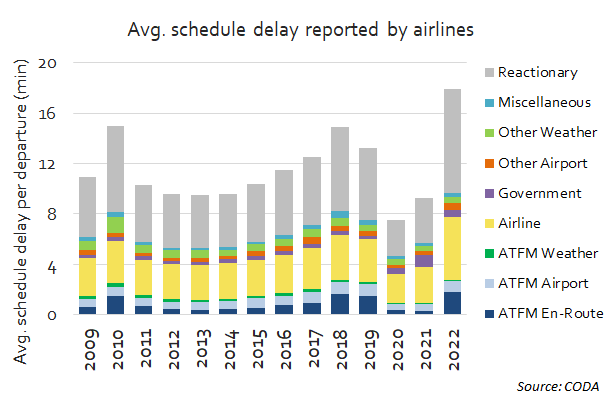}
  \caption{Scheduled delay causes for airlines, 2009-2022. (Source: PRR 2022.)}\label{fig:PRR_2022}
\end{figure}

The Performance Review Commission produces annual performance review reports (`PRRs') across a range of European air traffic management service delivery elements, often drawing on EUROCONTROL's Central Office for Delay Analysis (CODA) data. \autoref{fig:PRR_2022} shows the causes of average scheduled departure delay, as reported by airlines, in Europe, from 2009 to 2022 ~\citep{PRR_2022}. The picture is quite a complicated one in terms of interpreting year-on-year changes. This complexity has been compounded since the Covid-19 pandemic, in terms of new causal factors that need to be taken into account. However, two overall trends are clear. In each year, the major contributory factor is reactionary delay, and the second highest is airline-related. From 2009 - 2019 inclusive (i.e. pre-Covid-19), these averaged 27\% and 44\%, respectively (30\% and 44\%, respectively, in the year of the simulated data, 2014). This confirms the important and dominating role of these delay types and the need to model them effectively - an area where further research is certainly needed, both in the performance assessment domain \textit{per se} and beyond.

This leads to some concluding thoughts for this article as to when, in particular, one would want to apply a model of a model in the realm of ATM simulations. Of course, not all analyses for which simulations are used would benefit from metamodelling. For example, if the goal of simulation were to be to assess whether a certain new procedure respects given safety requirements, the precision, accuracy and eventual outlier events are of utmost importance, which would not be expected to lend itself to the application of metamodelling. However, when macromodelling the ATM system with the goal of performance assessment (for example), metamodelling offers the following benefits:

\begin{itemize}
    \item It facilitates more comprehensive analyses, due to the efficient utilisation of few simulations to generalise across scenarios, increasing the potential number of scenarios considered in a given amount of (modelling) time and strengthening the scalability of Solution assessment.
    \item It makes simulation results more explainable, facilitating interpretation, which can be of help to convey the results to policy makers (and other decision makers), especially those who are not simulation experts, and to improve predictability relating to potential operational and policy changes.   
    \item It can especially enhance scenario-based and `what-if' analyses, as it can be used to explore the scenario space of the simulator with higher efficiency, often including improved understanding of broad KPI interdependencies.
    \item Metamodelling is a broader \textit {enabler}. It facilitates and promotes a potential move towards a more fully consolidated performance assessment framework, ultimately accommodating a suite of simulators, each tailored to one or more KPAs or modelling capabilities (such as environmental impact, airspace structure, passenger itineraries); this, in turn, would support a future, truly common database of parameters (such as traffic assumptions, estimated aircraft performance, assumed fuel and carbon prices) and a common API syntax (one simulator being able to call elements of another), preferably in an open source environment - thus driving a more service-oriented architecture. 
\end{itemize}

\section*{Acknowledgements}
This work was supported by the NOSTROMO (Next-generation Open-Source Tools for ATM PeRfOrmance Modelling and Optimisation) project, framed in the scope of the SESAR 2020 Exploratory Research topic SESAR-ER4-26-2019, `ATM Validation for a Digitalised ATM,' with focus on the `Macro-modelling applied to Air Traffic Management' area and funded by SESAR Joint Undertaking through the European Union's Horizon 2020 research and innovation programme under grant agreement No 892517.


\section*{Authors Contribution}
The authors confirm their contribution to the paper as follows: study conception and design, methodology, software: C. Riis; data collection: C. Riis and G. Gurtner; analysis and interpretation of results: all; draft manuscript preparation: C. Riis, F. Antunes, T. Bolic., G. Gurtner, and A. Cook All authors reviewed the results and approved the final version of the manuscript.

\bibliographystyle{cas-model2-names}

\bibliography{bib}

\appendix

\section{XGBoost}\label{app:xgboost}
A Gradient Boosting Machine (GBM) is an ensemble of weak predictors used for regression and classifications in machine learning with great success \citep{nielsen2016tree}.
The model is built in a stage-wise fashion similar to other ensemble methods, such as  AdaBoost or Random Forest. We will consider the weak predictors to be a small regression tree $h_m(\bm{x};\bm{a}_m)$ with hyperparameters $\bm{a}_m$ for the input data point $\bm{x}$. Then the GBM is given as
\begin{equation}
F\left(\bm{x} ;\left\{\beta_{m}, \bm{a}_m\right\}_{m=1}^{M}\right)%
=\sum_{m=1}^{M} \beta_{m} h_{m}\left(\bm{x} ; \bm{a}_m\right),
\end{equation}
where $\beta_m$ is the learning rate and $M$ is the number of boosting steps. 
The optimal solution is then given by the function minimising the expected loss of all the data  $\mathcal{D}=(X, \bm{y})$ as
\begin{equation}
\centering
F^{*}=\arg \min _{F} \mathbb{E}_{X, \bm{y}}[L(\bm{y}, F(X ; \left\{\beta_{m}, \bm{a}_m\right\}_{m=1}^{M})],
\end{equation}
where $L(\bm{y}, F(X;\left\{\beta_{m}, \bm{a}_m\right\}_{m=1}^{M}))$ is the loss function, e.g. root mean square error (RMSE).

To achieve an efficient implementation of a gradient boosting machine (GBM) with decision trees, the XGBoost framework \citep{chen2016xgboost} is utilised. XGBoost employs a depth-wise tree growth approach and constructs trees up to a certain depth. This implementation benefits from the high speed of GBM by enabling the computation of the splits in parallel. Additionally, XGBoost incorporates several techniques to prevent overfitting, such as L2-regularisation and the learning rate as a shrinkage parameter.

\subsection{Optimising XGBoost}
In accordance with the standard practice in machine learning, we optimise the hyperparameters of each XGBoost model by employing the train-validation-test split technique to ensure strong generalisation performance across all relevant key performance indicators. We follow the setup of \cite{riis2022explainablemetamodel} and construct a training dataset comprising of 50,000 simulations, as well as a test dataset of 10,000 simulations, both generated through Latin hypercube sampling. Subsequently, we randomly divide the training dataset into a smaller training dataset of 40,000 simulations and a validation dataset of 10,000 simulations, which allows us to fine-tune the hyperparameters of XGBoost using a grid search. Through 10-fold cross-validation splits, we optimise the maximum depth of each decision tree, L2 regularisation, and learning rate by specifying a search space for each hyperparameter, as in \autoref{tab:grid_search}. 
\begin{table}[!ht]
    \centering
    \caption{Hyperparameters optimised with a grid search.}
    \label{tab:grid_search}
    \begin{tabular}{ll}
    \toprule
    \textbf{Hyperparameter} & \textbf{Space}\\
    \midrule
    Max depth     & [3, 4, 5, 6, 7, 8, 9] \\
    L2-regularisation     & [0, 0.1, 0.2, 0.3, 0.4, 0.5]\\
    Learning rate & [0.01, 0.05, 0.1, 0.2, 0.3, 0.4]\\
    \bottomrule
    \end{tabular}
\end{table}
Additionally, we employ the exact greedy algorithm to select the decision tree splits. We imposed a maximum limit of 1000 boosting iterations for all models and early stopping was employed if the validation accuracy did not improve over a span of five boosting steps. The optimal hyperparameters are listed in \autoref{tab:hyperparameters_XGBoost}.
\begin{table}[!ht]
    \centering
    \caption{Hyperparameters for XGBoost. The hyperparameters have been found using the grid search technique.} 
    \label{tab:hyperparameters_XGBoost}
    \begin{tabular}{lrccc}
    \toprule
    \textbf{KPI} & \textbf{\#sim} & \textbf{Max depth} & \textbf{Learning rate} & \textbf{L2-reg.}\\ 
    \midrule
    {{Flight arrival delay}}  
          & 50000        & 3 & 0.01 & 0.5 \\
          & 1000        &  3 & 0.05 & 0.2 \\
    \midrule
    {{Flight departure delay}}
           & 50000        & 3 & 0.01 & 0.4 \\
           & 50000        & 3 & 0.10 & 0.0 \\
    \midrule
    {{Passenger arrival delay}}
           & 50000        &  5 & 0.01 & 0.5 \\
           & 1000        &  4 & 0.01 & 0.5 \\
    \midrule
    {{Planned absorbed delay}} 
          & 50000        & 4 & 0.05 & 0.2 \\
          & 1000        & 4 & 0.30 & 0.1 \\
    \midrule
    {{Holding time}}  
          & 50000        &3 & 0.20 & 0.5 \\
          & 1000        & 3 & 0.20 & 0.0 \\
    \midrule
    {{Fuel cost}}  
             & 50000        &   4 & 0.01 & 0.4 \\
          & 1000        &  5 & 0.01 & 0.0 \\
    \bottomrule
    \end{tabular}
\end{table}

\section{The effect of turnaround time on flight arrival and departure delay}\label{ap:tat_delays}
\begin{figure}[!ht]
    \centering
    \includegraphics[width=0.48\textwidth]{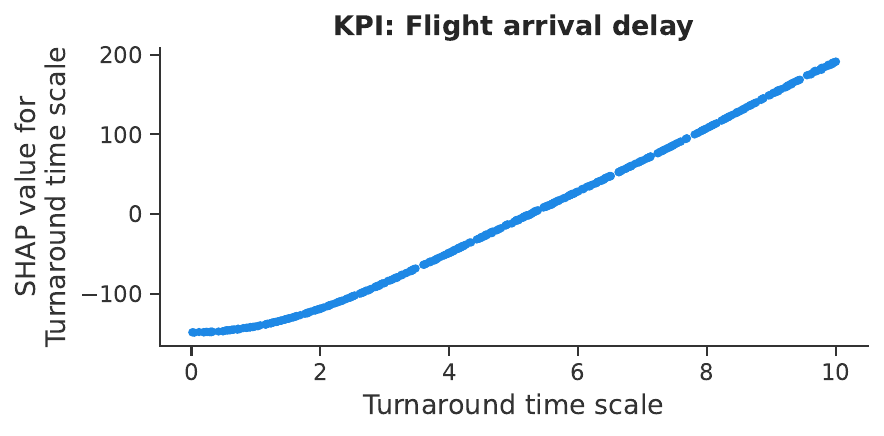}
    \includegraphics[width=0.48\textwidth]{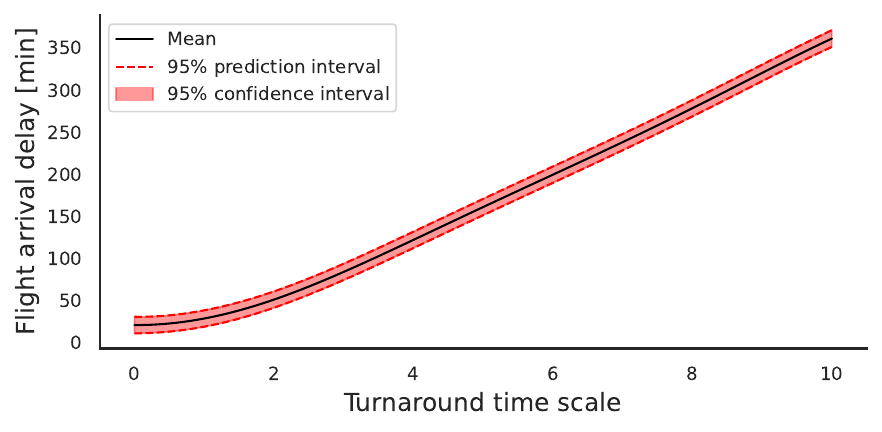}
    \caption{The impact of the turnaround time scale on flight arrival delay. \textbf{Left}: SHAP values. \textbf{Right}: Model predictions.}
    \label{fig:flight_arrival_delay}
\end{figure}

\begin{figure}[!ht]
    \centering
    \includegraphics[width=0.48\textwidth]{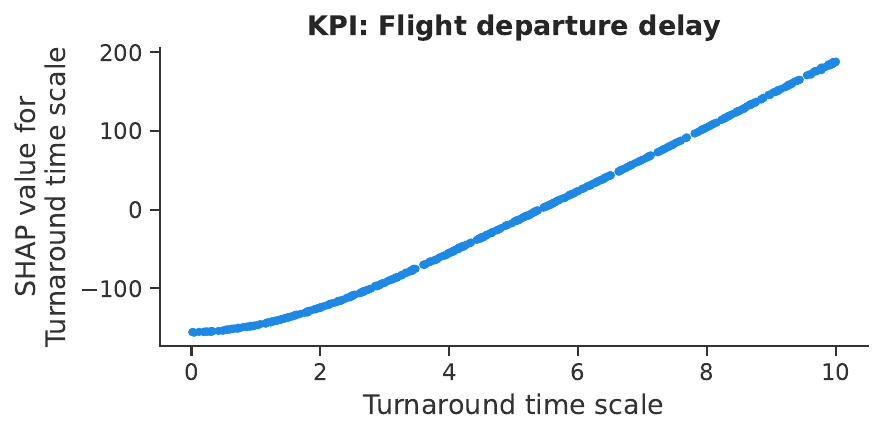}
    \includegraphics[width=0.48\textwidth]{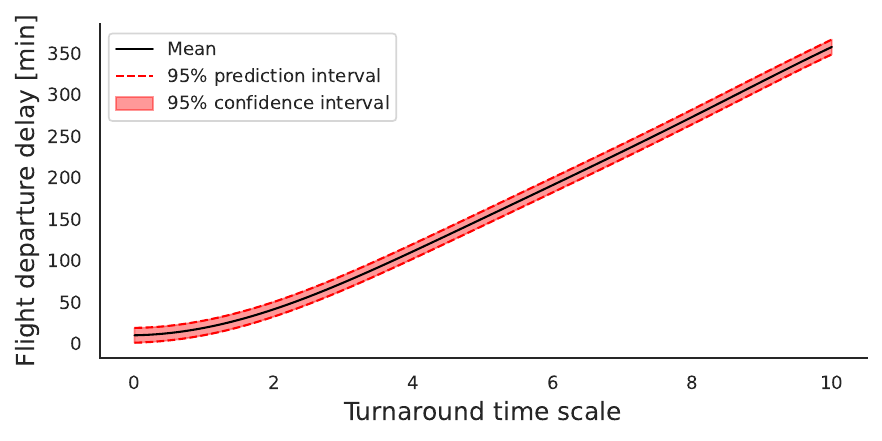}\\
    \caption{The impact of the turnaround time scale on flight departure delay. \textbf{Left}: SHAP values. \textbf{Right}: Model predictions.}
    \label{fig:flight_departure_delay}
\end{figure}

The SHAP values and model predictions for the flight arrival and departure delay are very similar, cf. \autoref{fig:flight_arrival_delay} and \autoref{fig:flight_departure_delay}. In the following, we describe the arrival delay.
In the left plot in \autoref{fig:flight_arrival_delay}, we see an almost linear relationship between the input and the output, and for example that if the turnaround time is near 0, the delay is the base value subtracted by 150 min. On the other hand, a turnaround time close to 10 gives a delay at 200 min above the mean. The right plot in \autoref{fig:flight_arrival_delay} shows the prediction of the delays, when the features are set to the default values, and only the turnaround time scale is changed. The predictions confirms the trend of the SHAP values, and we see that there is only little uncertainty about the mean predictions, and almost no stochasticity in the delays.

\section{Active learning curves with stopping criterion}\label{app:minimize_variance}
\begin{figure}[!ht]
    \centering
    \includegraphics[width=\textwidth]{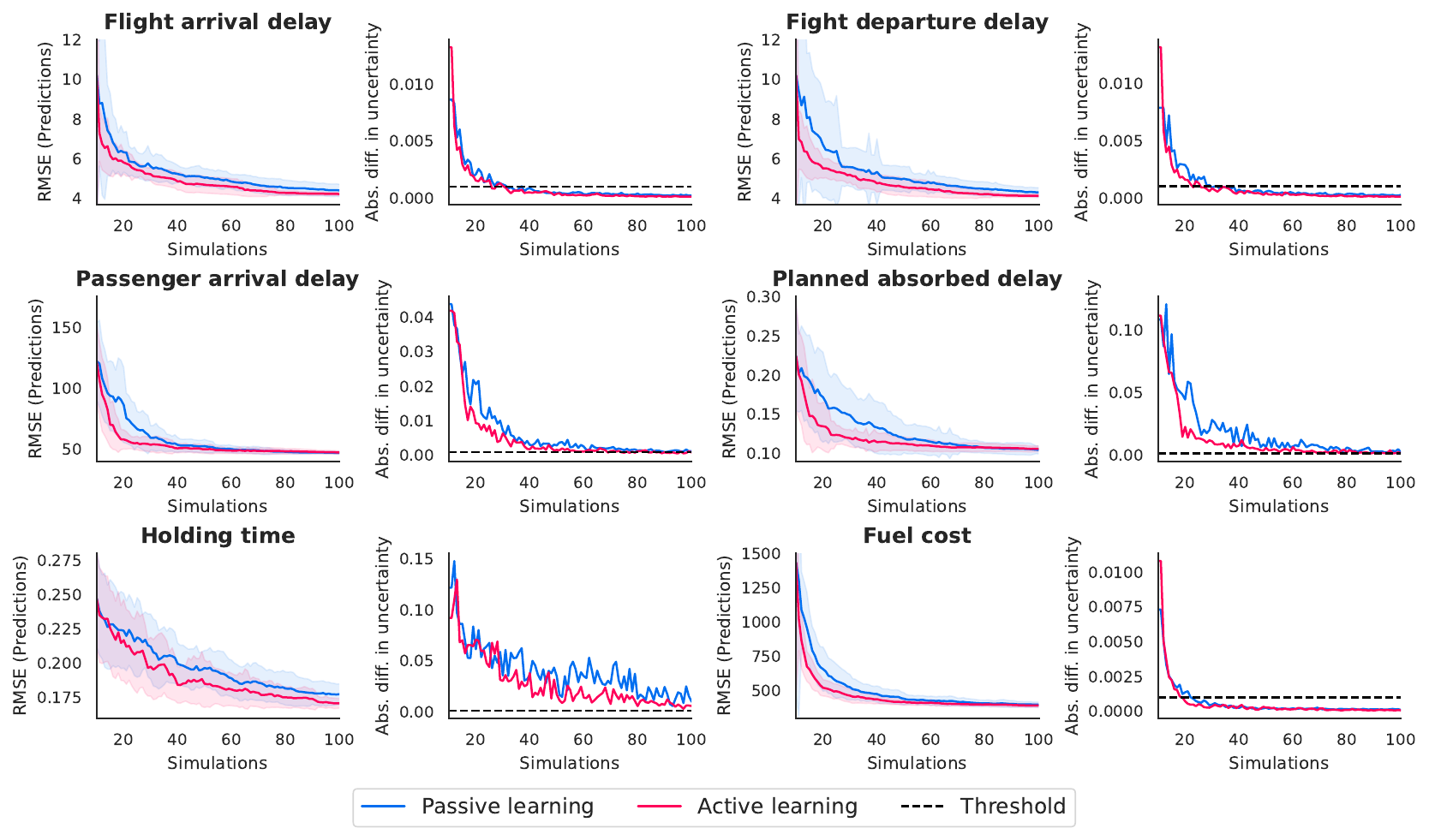}
    \caption{Active learning curves for the six KPIs. The figure shows the mean results of 30 repetitions, represented by lines, with the shaded area indicating the range of $\mu \pm \sigma$. For each KPI, the left panel shows the root mean square error (RMSE) of the predictions for the two acquisition functions over 90 acquisitions, and the right panel displays the absolute difference in the epistemic uncertainty between the active learning iterations, where the epistemic
uncertainty for the model is predicted for the unlabelled data set.}
    \label{fig:minimize_variance_all}
\end{figure}

In \autoref{fig:minimize_variance_all}, we show two plots for each key performance indicator. To the left, we show the active learning loss curves, and to the right, we show the absolute difference in the epistemic uncertainty between the active learning iterations, where the epistemic uncertainty for the model is predicted for the unlabelled data set. Comparing the right plots to the left plots\autoref{fig:minimize_variance_arrdel}, we see that curves for the absolute difference in the uncertainty approximately follow that of the RMSE of the SHAP values. 

\FloatBarrier

\end{document}